\documentclass[sigconf]{acmart}
\usepackage{graphicx} %
\usepackage{amsmath}

\usepackage{amssymb}
\usepackage{color}
\usepackage{xspace}
\usepackage{hyperref}
\usepackage{comment}
\usepackage{enumitem}
\usepackage{multirow}
\usepackage{diagbox}
\setlist[itemize]{leftmargin=*}
\usepackage{xcolor}
\usepackage{caption}
\usepackage{tabularx}
\usepackage{makecell}
\usepackage[pagebackref=true,breaklinks=true,colorlinks,bookmarks=false]{}
\usepackage{hyperref}
\hypersetup{
  colorlinks,
  linkcolor={blue!70!green},
  citecolor={green!70!blue},
  urlcolor={black}
}

\newcommand{\sysname}{\ensuremath{\mathsf{ArtistAuditor}}\xspace}
\newcommand{\etal}{\textit{et al.}\xspace}
\newcommand{\ie}{\textit{i.e.}\xspace}
\newcommand{\eg}{\textit{e.g.}\xspace}

\newcommand{\mypara}[1]{\smallskip\noindent\textbf{#1.} \xspace}

\usepackage{hyperref}
\makeatletter
\def\UrlAlphabet{%
      \do\a\do\b\do\c\do\d\do\e\do\f\do\g\do\h\do\i\do\j%
      \do\k\do\l\do\m\do\n\do\o\do\p\do\q\do\r\do\s\do\t%
      \do\u\do\v\do\w\do\x\do\y\do\z\do\A\do\B\do\C\do\D%
      \do\E\do\F\do\G\do\H\do\I\do\J\do\K\do\L\do\M\do\N%
      \do\O\do\P\do\Q\do\R\do\S\do\T\do\U\do\V\do\W\do\X%
      \do\Y\do\Z}
\def\UrlDigits{\do\1\do\2\do\3\do\4\do\5\do\6\do\7\do\8\do\9\do\0}
\g@addto@macro{\UrlBreaks}{\UrlOrds}
\g@addto@macro{\UrlBreaks}{\UrlAlphabet}
\g@addto@macro{\UrlBreaks}{\UrlDigits}
\makeatother

\AtBeginDocument{%
  }

\copyrightyear{2025}
\acmYear{2025}
\setcopyright{acmlicensed}\acmConference[WWW '25]{Proceedings of the ACM Web Conference 2025}{April 28-May 2, 2025}{Sydney, NSW, Australia}
\acmBooktitle{Proceedings of the ACM Web Conference 2025 (WWW '25), April 28-May 2, 2025, Sydney, NSW, Australia}
\acmDOI{10.1145/3696410.3714602}
\acmISBN{979-8-4007-1274-6/25/04}

\begin{document}

\title{\sysname: Auditing Artist Style Pirate in Text-to-Image Generation Models}

\author{Linkang Du}
\orcid{0009-0004-9028-9326}
\authornote{Both authors contributed equally to this research.}
\affiliation{%
  \institution{Xi'an Jiaotong University}
  \city{Xi'an}
  \country{China}}
\email{linkangd@xjtu.edu.cn}

\author{Zheng Zhu}
\orcid{0009-0009-5268-9667}
\authornotemark[1]
\affiliation{%
  \institution{Zhejiang University}
  \city{Hangzhou}
  \country{China}
}
\affiliation{%
  \institution{The Chinese University of Hong Kong}
  \city{Hong Kong}
  \country{China}
}
\email{zjuzhuzheng@zju.edu.cn}

\author{Min Chen}
\orcid{0000-0002-1128-7989}
\affiliation{%
  \institution{Vrije Universiteit Amsterdam}
  \city{Amsterdam}
  \country{Netherlands}
}
\email{m.chen2@vu.nl}

\author{Zhou Su}
\orcid{0000-0002-2875-0458}
\affiliation{%
  \institution{Xi'an Jiaotong University}
  \city{Xi'an}
  \country{China}
}
\email{zhousu@ieee.org}

\author{Shouling Ji}
\orcid{0000-0003-4268-372X}
\affiliation{%
  \institution{Zhejiang University}
  \city{Hangzhou}
  \country{China}
}
\email{sji@zju.edu.cn}

\author{Peng Cheng}
\orcid{0000-0002-4221-2162}
\affiliation{%
  \institution{Zhejiang University}
  \city{Hangzhou}
  \country{China}
}
\email{lunarheart@zju.edu.cn}

\author{Jiming Chen}
\orcid{0000-0003-3155-3145}
\affiliation{%
  \institution{Zhejiang University}
  \city{Hangzhou}
  \country{China}
}
\affiliation{%
  \institution{Hangzhou Dianzi University}
  \city{Hangzhou}
  \country{China}
}
\email{cjm@zju.edu.cn}

\author{Zhikun Zhang}
\orcid{0000-0001-7208-3392}
\authornote{Zhikun Zhang is the corresponding author.}
\affiliation{%
  \institution{Zhejiang University}
  \city{Hangzhou}
  \country{China}
}
\email{zhikun@zju.edu.cn}
\renewcommand{\shortauthors}{Linkang Du et al.}

\begin{abstract}
Text-to-image models based on diffusion processes, such as DALL-E, Stable Diffusion, and Midjourney, are capable of transforming texts into detailed images and have widespread applications in art and design. 
As such, amateur users can easily imitate professional-level paintings by collecting an artist's work and fine-tuning the model, leading to concerns about artworks' copyright infringement. 
To tackle these issues, previous studies either add visually imperceptible perturbation to the artwork to change its underlying styles (perturbation-based methods) or embed post-training detectable watermarks in the artwork (watermark-based methods).
However, when the artwork or the model has been published online, \ie, modification to the original artwork or model retraining is not feasible, these strategies might not be viable. 

To this end, we propose a novel method for data-use auditing in the text-to-image generation model. 
The general idea of \sysname is to identify if a suspicious model has been fine-tuned using the artworks of specific artists by analyzing the features related to the style. 
Concretely, \sysname employs a style extractor to obtain the multi-granularity style representations and treats artworks as samplings of an artist's style. 
Then, \sysname queries a trained discriminator to gain the auditing decisions. 
The experimental results on six combinations of models and datasets show that \sysname can achieve high AUC values (>~0.937). 
By studying \sysname's transferability and core modules, we provide valuable insights into the practical implementation. 
Finally, we demonstrate the effectiveness of \sysname in real-world cases by an online platform Scenario.\footnote{\url{https://www.scenario.com/}} 
\sysname is open-sourced at \url{https://github.com/Jozenn/ArtistAuditor}. 
\end{abstract}



\maketitle

\section{Introduction}
\label{sec:introduction}
Text-to-image models represent a groundbreaking advancement in generative artificial intelligence (GAI), such as DALL-E~\cite{ramesh2021zeroshot}, Stable Diffusion~\cite{Rombach_2022_CVPR}, and Midjourney~\cite{ivanenko2022midjourney}, which can generate realistic images from textual descriptions. 
These models typically function by gradually refining a random pattern of pixels into a coherent image that matches the text, making them suitable for a variety of creative and practical applications~\cite{c-ai-cover, children-book, aigame, Li2021SRDiffSI, meng2023sepanner, wang2023k, chen2023quote, liu2023chatgpt}. 

\mypara{Relevance to the Web and the Security and Privacy Track}
These models are rapidly gaining popularity among users through web platforms due to their impressive capabilities, including open API interfaces and open-source implementations. 
For example, Midjourney receives around 32 million pageviews per day at around 7.5 pageviews per visit~\cite{midjourneystat23}. 
With the rapid development of text-to-image models, a user with little painting experience can use prompts to generate artwork at a professional level. 
As one of the sensational events, Jason M. Allen created his digital artwork with Midjourney and took first place in the digital category at the Colorado State Fair~\cite{winaward}. 
Recently, many platforms allow users to upload artworks and train the models that can generate artworks of similar style~\cite{civitai2022, aigame, MHM2023Finetune}. 
The ease of generating artwork using GAI might devalue the skill and expression involved in human-made artwork, diminishing the appreciation of human creativity. 
For instance, the artists feel that their unique styles are being appropriated when the market is flooded with AI-mimicked artworks~\cite{shan2023glaze}. 
This raises questions about dataset infringement, highly relevant to ``security and privacy of machine learning and AI applications.'' 

\mypara{Existing Solutions}
To protect the intellectual property (IP) of artists, a series of strategies have been proposed~\cite{CZWBZ23, Van_Le_2023_ICCV, shan2023glaze, Chen2023EditShieldPU, Zhao2023CanPP, Cui2023DiffusionShieldAW, Luo2023StealMA, WMZCZFWZC24, DCSJCCZ24}. 
The existing solutions can be classified into two categories by the underlying technologies, \ie, the perturbation-based methods~\cite{Chen2023EditShieldPU, shan2023glaze, Van_Le_2023_ICCV, Zhao2023CanPP} and the watermark-based methods~\cite{wang2023making, Cui2023DiffusionShieldAW, liu2024physics, Ma2023GenerativeWA, zhu2023detection}. 
The perturbation-based methods introduce subtle perturbations that alter the latent representation in the diffusion process, causing models to be unable to generate images as expected. 
The watermark-based methods inject imperceptible watermarks into artworks before they are shared. 
The diffusion model collects and learns the watermarked artworks. 
The artists can then validate the infringements by checking if the watermarks exist in the generated images. 
Membership inference (MI)~\cite{Shokri2016MembershipIA, Chen2019GANLeaksAT, CZWBZ23, chen2024janus} is another technique to determine whether specific data was used to train or fine-tune the diffusion model~\cite{Wang2024PropertyEI, Kandpal2023UserIA, Pang2023BlackboxMI, Duan2023AreDM}. 

However, previous studies face several limitations. 
First, both the perturbation-based and the watermark-based methods need to manipulate the original images, \ie, injecting perturbation or watermark, thus compromising data fidelity. 
The perturbation may also diminish the model's generation quality. 
Second, perturbation-based and watermark-based strategies require retraining the model to be effective. 
Thus, they may not suit the model already posted online. 
For the MI methods, the existing approaches~\cite{Pang2023WhiteboxMI, Duan2023AreDM, Kong2023AnEM, Fu2023APF, Hu2023MembershipIO, Matsumoto2023MembershipIA} for diffusion models usually require the access to structure or weights of the model, which limits their applicability in black-box auditing scenarios. 
Although some MI strategies target the black-box settings~\cite{Zhang2023GeneratedDA, Kandpal2023UserIA, Wang2024PropertyEI, Pang2023BlackboxMI, DCSJCCZ24, DZCJCCZ24}, they are not well suited to our auditing task. 
We will go depth in~\autoref{sec:discussion} and compare them with \sysname in~\autoref{sec:evaluation}. 

\begin{figure}[!t]
\centering
\includegraphics[width=0.86\hsize]{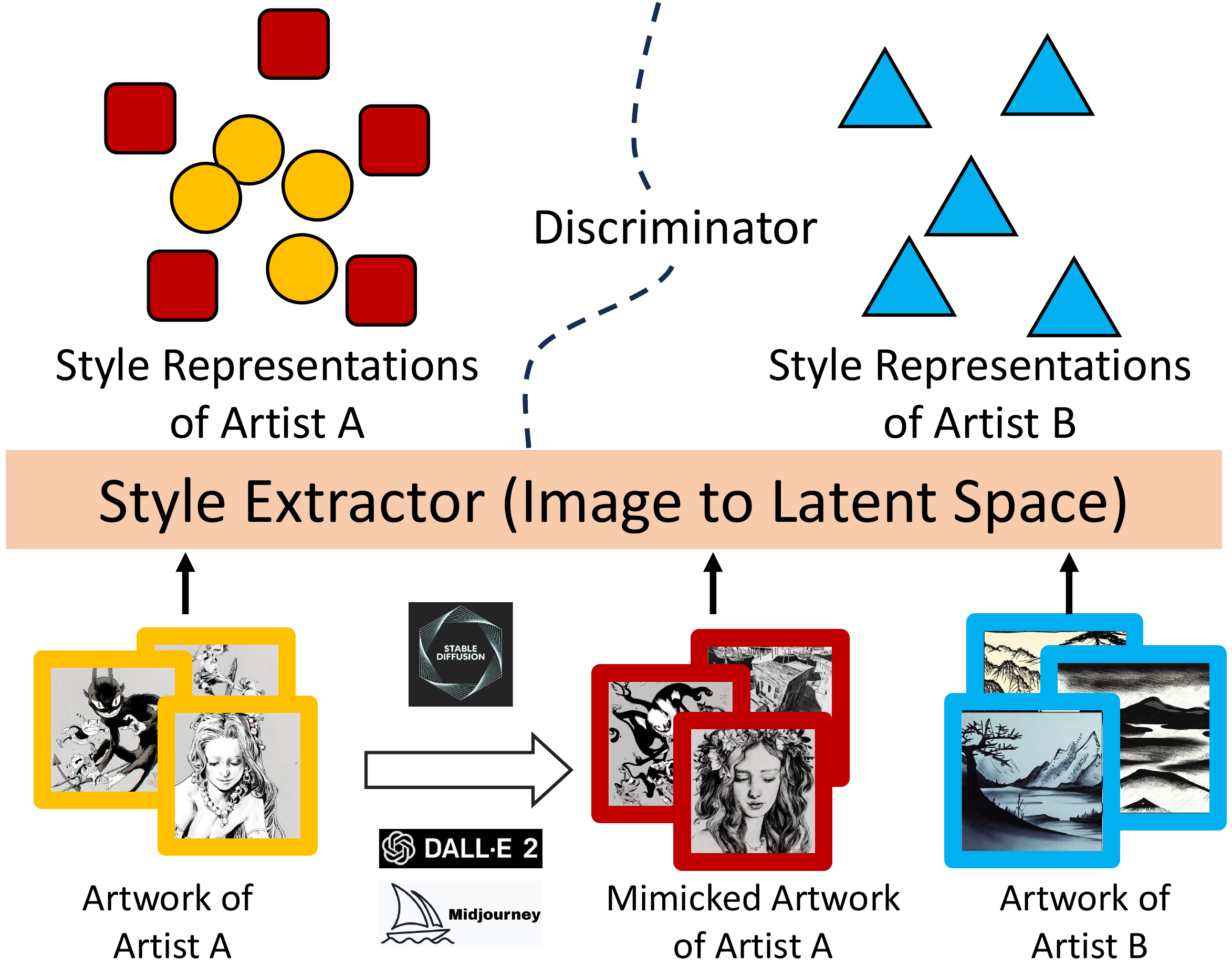}
\caption{Intuitive explanation of \sysname. Images with orange borders represent artist A's artworks, red borders indicate artworks mimicked by models, and blue borders show B's artworks.
The discriminator identifies the style pirate based on the latent representations of the artworks.}
\label{fig:intuition}
\vspace{-0.4cm}
\end{figure}

\mypara{Our Proposal}
In this paper, we propose a novel artwork copyright auditing method for the text-to-image models, called \sysname, which can identify data-use infringement without sacrificing the artwork's fidelity. 
We are inspired by the fact that artworks within an artist's style share some commonality in latent space. 
Thus, the auditor can mine the style-related features in an artist's works to form the auditing basis. 
\autoref{fig:intuition} provides a schematic diagram of \sysname, where the core components are the style extractor and discriminator. 
Since the entire feature space retains a variety of information about the artwork (\eg, objects, locations, and color), the auditor needs to extract the style-related features at different levels of granularity. 
The auditor then adopts a discriminator to predict the conference score. 
The discriminator outputs a positive result if the generated images closely match the style of the artist; otherwise, it outputs a negative prediction. 
Finally, we leverage two strategies to process the confidence scores and derive the decision. 

\mypara{Evaluation}
Our experimental results on three popular diffusion models (Stable Diffusion v2.1~\cite{stable2-1}, Stable Diffusion XL~\cite{Podell2023SDXLIL}, and Kandinsky~\cite{Razzhigaev2023KandinskyAI}) and two artistic datasets (Wikiart~\cite{artgan2018} and self-collected dataset) consistently achieve AUC values of \sysname above 0.937. 
By comparing original artworks with mimicked ones, we find that \sysname can accurately identify imitations that differ in content from the originals but pirate the artist's style. 
In addition, we evaluate four influential factors from two aspects for the practical adoption of \sysname. 
The first aspect focuses on the transferability of \sysname. 
In practice, the auditor is unaware of the selected artworks or the image captioning model used to fine-tune the suspicious model. 
Thus, we assess the transferability of \sysname between datasets and models.
When the selected artworks are disjoint with those to fine-tune the suspicious model, the auditing accuracy of \sysname only drops by 2.6\% compared to the complete overlap scenario on the Kandinsky model. 
For different captioning models, \sysname can still maintain an accuracy of 85.3\% and a false positive rate below 13.3\%. 
The second aspect focuses on the core modules of \sysname, namely data augmentation and distortion calibration. 
Data augmentation aims to increase the number of artworks available for training discriminators. 
Distortion calibration is used to mitigate the negative impact on auditing accuracy of potential stylistic distortions in the generation process. 
The results demonstrate that both modules enhance the accuracy of \sysname in most experimental settings. 
Finally, we show the effectiveness of \sysname in real-world cases by a commercial platform \href{https://www.scenario.com/}{Scenario}. 

\mypara{Contributions}
Our contributions are three-fold:
\begin{itemize}
    \item To our knowledge, \sysname is the first dataset auditing method to use multi-granularity style representations as an intrinsic fingerprint of the artist.
    \sysname is an efficient and scalable solution, using under 13.18 GB of GPU memory per artist and enabling parallel auditing due to decoupling processes among artists.

    \item We show the effectiveness of \sysname on three mainstream diffusion models. By systematically evaluating \sysname from several aspects, \ie, the dataset transferability, the model transferability, and the impact of the different modules, 
    we summarize some useful guidelines for adopting \sysname in practice. 
    \item By implementing \sysname on the online model fine-tuning platform \href{https://www.scenario.com/}{Scenario}, we show that \sysname can serve as a potent auditing solution in real-world text-to-image scenarios. 
\end{itemize}

\subsection{Ethical Use of Data and Informed Consent}
We strictly followed ethical guidelines by using publicly available, open-source datasets and models under licenses permitting research and educational use. 
As these datasets were curated and released by third parties, direct informed consent was not applicable. 
However, we are committed to ethical data use and will comply with all licensing terms for any future modifications or redistribution. 

\section{Background}
\label{sec:background}
\subsection{Text-to-Image Generation}
Generative adversarial network (GAN)~\cite{Goodfellow2014GenerativeAN, Creswell2017GenerativeAN, Karras2021AliasFreeGA} and diffusion model (DM)~\cite{ramesh2021zeroshot, Rombach_2022_CVPR, ivanenko2022midjourney} have been used in text-to-image tasks. 
GAN in this space might struggle with the fidelity and diversity of the images. 
Inspired by the physical process of diffusion, where particles spread over time, DM represents a significant development in generative models. 
These models function through a two-phase process: a forward process that gradually adds noise to an image over a series of steps until it becomes random noise and a reverse process where the model learns to reverse this, reconstructing the image from noise. 
The forward process gradually adds noise to an image $x_0$ over a series of steps $T$. 
This process can be represented as a Markov chain, where each step adds Gaussian noise. 

\begin{small}
\begin{equation}
x_t = \sqrt{\alpha_t} x_{t-1} + \sqrt{1 - \alpha_t} \epsilon_t , 
\end{equation}
\end{small}

\noindent
where $x_t$ is the noisy image at step $t$, $x_{t-1}$ is the image from the previous step, $\epsilon_t$ is the noise added at step $t$ sampled from a normal distribution, \ie, $\epsilon_t \sim \mathcal{N}(0, I)$. 
$\alpha_t$ is a variance schedule determining how much noise to add at each step. 
It's a predefined sequence of numbers between 0 and 1. 

The model learns to generate images by reversing the noise addition in the reverse process. 
At step $t$, the model predicts the noise $\epsilon_t$ added in the forward process and then uses this to compute the previous step's image $x_{t-1}$. 

\begin{small}
\begin{equation}
x_{t-1} = \frac{1}{\sqrt{\alpha_t}} \left( x_t - \frac{1 - \alpha_t}{\sqrt{1 - \bar{\alpha}_t}} \epsilon_{\theta}(x_t, t) \right), 
\end{equation}
\end{small}

\noindent
where $\epsilon_{\theta}(x_t, t)$ is the noise predicted by the model (parameterized by $\theta$), given $x_t$ and the time step $t$. 
$\bar{\alpha}_t$ is the cumulative product of $\alpha_i$ up to step $t$, \ie, $\bar{\alpha}_t = \prod_{i=1}^{t} \alpha_i$. 
The model starts with a sample of pure noise $x_T \sim \mathcal{N}(0, I)$ and applies this denoising step iteratively to arrive at a generated data point $x_0$. 
The model training involves learning the parameters $\theta$ to accurately predict the noise $\epsilon_t$ at each step.
Diffusion models excel at generating highly detailed and coherent images, showing great flexibility and stability in training.

\subsection{Style Piracy}
\label{sec:style-mimicry}

\begin{figure}[!t]
\centering
\includegraphics[width=\hsize]{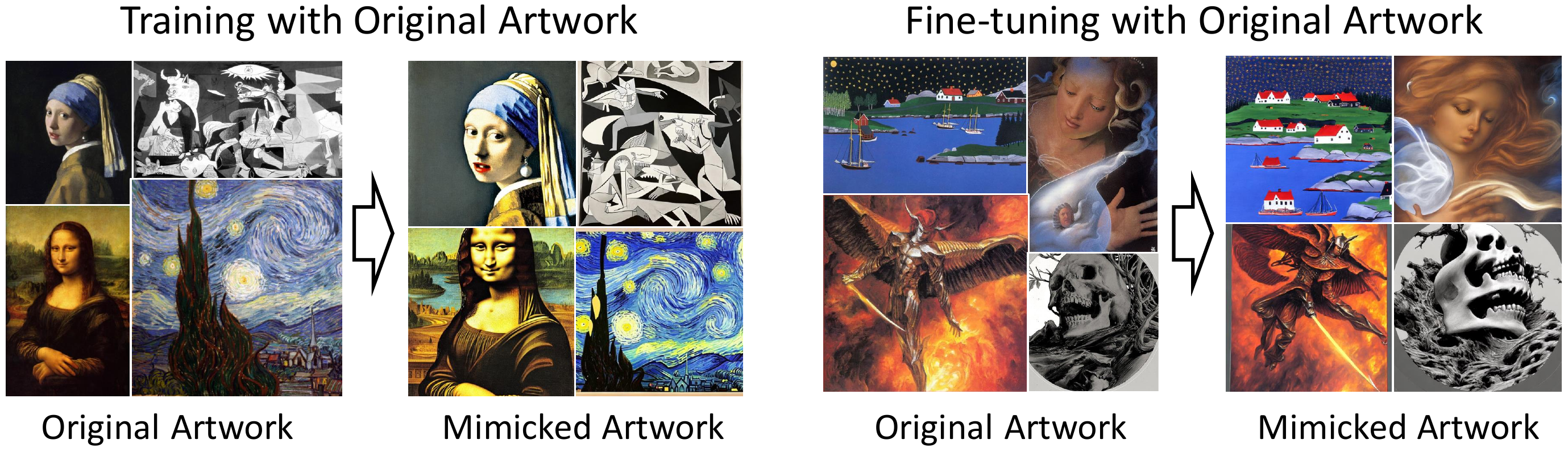}
\caption{An example of stylistic imitation by Stable Diffusion. 
Left: original artwork. 
Right: generated artwork. 
}
\label{fig:style-mimicry}
\vspace{-0.4cm}
\end{figure}

\mypara{Technique} 
The concept of style piracy in the text-to-image field refers to using diffusion models to create images that closely resemble a specific artistic style. 
The first way is to train the diffusion models from scratch on a large dataset of images that includes the target artist's artworks.
It allows the model to learn and replicate the artist's style. 
A simple style piracy directly queries a text-to-image model using the artist's name. 
For instance, on the left of \autoref{fig:style-mimicry}, we utilize Stable Diffusion to imitate the style of artworks. 

However, since the huge overhead for training the diffusion models, the adversary tends to fine-tune diffusion models for style piracy, \ie, adjusting the diffusion models by a small set of the target artist's artwork~\cite{Gal2022Image, Hu2021Lora, kumari2023multi, ruiz2023dreambooth}. 
This dataset encompasses unique elements like specific brushwork, color schemes, and compositional techniques characteristic of the artist's style. 
The fine-tuning process involves continuous learning and adjustment to enhance the model's ability to apply these style characteristics accurately to various contents. 
On the right of \autoref{fig:style-mimicry}, we demonstrate the model's imitation ability after fine-tuning. 

\section{Problem Statement}
\label{sec:preliminaries}
\subsection{System and Threat Model}
\label{sec:threat-model}
\begin{figure}[!t]
\centering
\includegraphics[width=0.9\hsize]{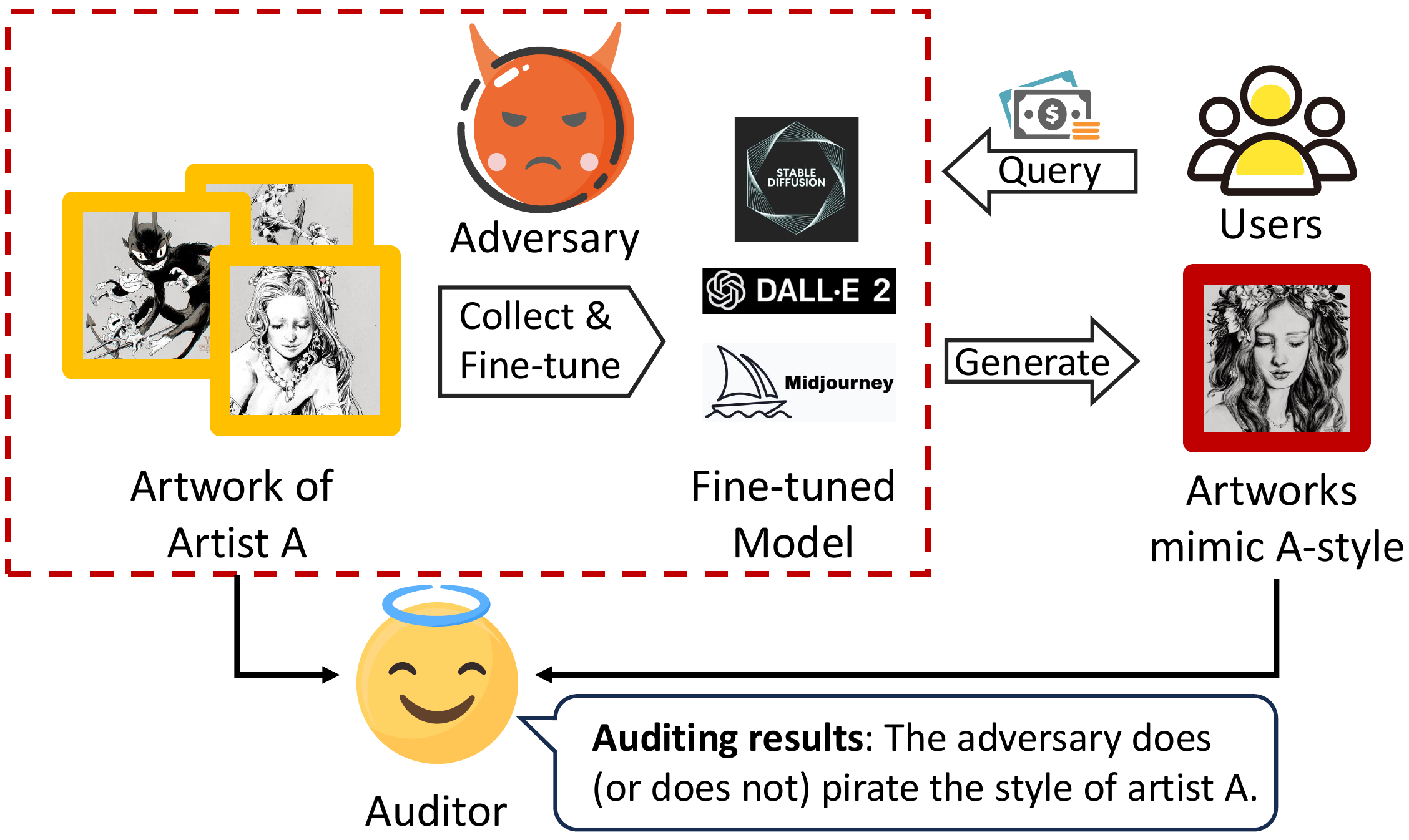}
\caption{An example of the application scenario. 
The auditor acquires the auditing results by comparing the style representations between the original artwork of artist A and the artworks generated by the fine-tuned model. }
\label{fig:application-scenario}
\vspace{-0.4cm}
\end{figure}

\mypara{Application Scenarios}
Comparing training the diffusion models from scratch, the adversary can easily implement style piracy by fine-tuning the models. 
Thus, we mainly consider the fine-tuning scenarios in this work, where the adversary collects a small set of artworks from an artist and adjusts the models' parameters to mimic the artist's style. 
\autoref{fig:application-scenario} illustrates a typical application case. 
Since many artists post their works online, adversaries can easily collect them by searching the artist's name. 
They fine-tune the diffusion model to generate artwork miming the artist's style. 
The artist stumbles upon the model's ability to generate artwork similar to his/her style and thus suspects the model's unauthorized use of his/her artwork for fine-tuning. 
The artist adopts \sysname to audit the suspicious model. 

\mypara{Auditor's Background Knowledge and Capability}
For the above application scenarios, we consider the auditor to have black-box access to the suspicious text-to-image model. 
During the auditing, the auditor can access the artist's artworks and use a low-end consumer GPU to extract the style representations. 
Additionally, the auditor does not have prior knowledge of the selected artworks by the adversary. 
Note that this is the most general and challenging scenario for the auditor. 
The auditor can collect the generated images by querying the suspicious model with legitimate prompts. 

\subsection{Design Challenges}
\label{sec:design-challenges}
From the above analysis, we face two challenges during the design of the data-use auditing method for text-to-image models.
The primary obstacle lies in the absence of a mathematical framework to precisely define and quantify ``artistic styles''. 
Generally, the style of an artist is defined by a multifaceted combination of elements, each contributing to its unique aesthetic and thematic identity. 
For instance, Claude Monet is regarded as the quintessential impressionist. 
Monet's work is characterized by his fascination with light and its effects on the natural world. 
Edgar Degas is also considered an impressionist, and his style differs from that of Monet. 

The second challenge is that the diffusion models often are fine-tuned with a set of artworks from multiple artists. 
This causes the features of these artists' artworks to interact, interfering with the effectiveness of auditing for a specific artist. 
Thus, the proposed method must effectively extract the unique features of an artist's artworks from the generated content to make accurate judgments. 

\begin{figure*}[!t]
\centering
\includegraphics[width=\hsize]{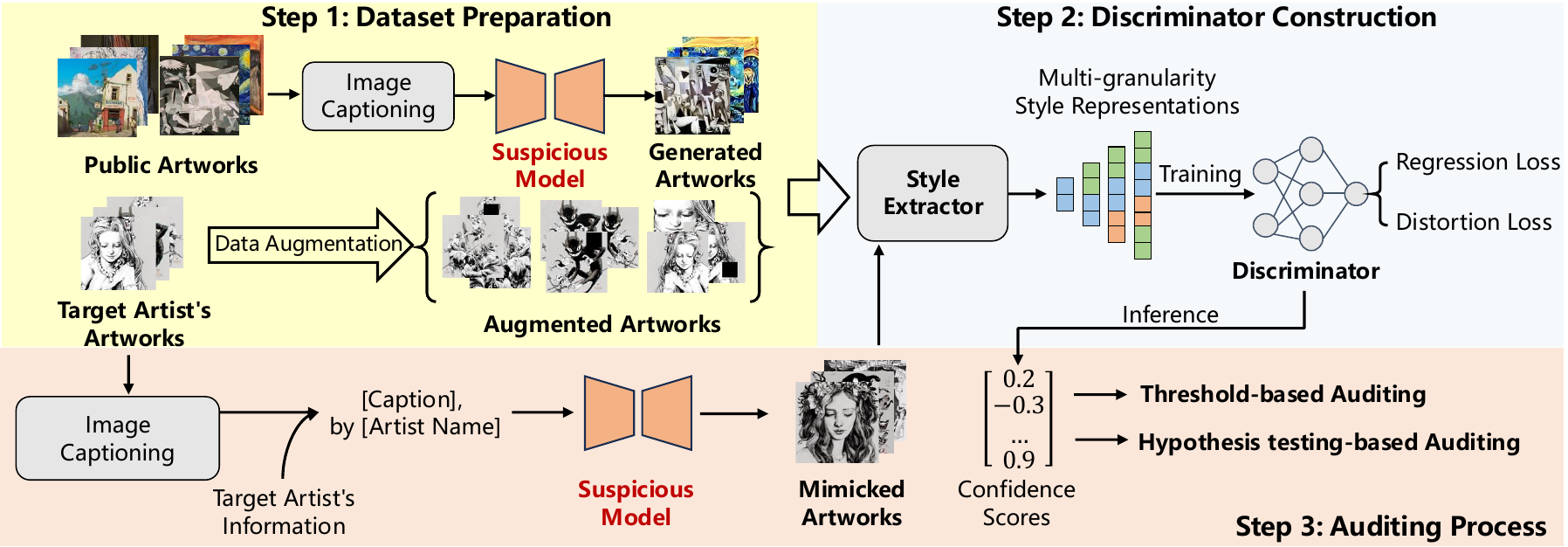}
\caption{The workflow of \sysname contains three steps, \ie, dataset preparation, discriminator construction, and auditing process. 
\sysname first collects the public artworks and generated artworks by the suspicious model, then extracts the multi-granularity style representations to train the discriminator. 
Finally, \sysname extracts the style features of mimicked artworks and makes the auditing decisions based on the outputs of the discriminator. 
}
\label{fig:framework}
\end{figure*}

\section{Methodology}
\subsection{Intuition}
Inspired by ~\cite{gatys2015neural, zhang2022domain}, 
we leverage latent representations at different layers of convolutional neural networks (CNNs) as the fingerprint of the artist's style. 
In CNNs, initial layers typically capture low-level features such as edges, colors, and textures, \ie, more closely related to the concrete elements of artworks. 
The deeper layers capture higher-level features, which represent more abstract information, like object parts or complex shapes. 
Then, we resort to a regression model to compress these style representations into a set of confidence scores to make the final auditing decision. 

\subsection{Workflow of \sysname} 
\label{sec:workflow-of-auditor}
For clarity, an artist whose artworks are being audited is called \textit{target artist}. 
If the suspicious model is fine-tuned on the target artist's artwork, the discriminator should output a positive auditing result for it; otherwise, a negative auditing result. 
\autoref{fig:framework} illustrates the workflow of \sysname. 

\mypara{Step 1: \underline{D}ataset \underline{P}reparation (DP)}
The first step collects three types of artworks, \ie, public artworks, generated artworks, and augmented artworks. 
The public artworks are the world-famous images published online, which are commonly included in the pre-training of the diffusion model~\cite{Rombach_2022_CVPR, schuhmann2022laion}, such as the paintings of Picasso and Da Vinci. 
Based on these public artworks, the auditor can create a set of prompts to query the suspicious model and obtain their mimicked version. 
Specifically, we adopt the CLIP interrogator\footnote{https://github.com/pharmapsychotic/clip-interrogator?tab=readme-ov-file} to generate the caption for each public artwork. 
Then, we take these captions as prompts to query the suspicious model and get the mimicked artworks of these world-famous artists. 
Since the artworks of the target artist may be insufficient to train the discriminator, we utilize data augmentation to expand the number of artworks and gain the augmented artworks.
We adopt the popularly used random cropping, random horizontal flipping, random cutouts, Gaussian noise~\cite{Cohen2019CertifiedAR}, impulse noise~\cite{Koli2013LiteratureSO}, and color jittering~\cite{Krizhevsky2012ImageNetCW}, in existing works~\cite{Szegedy2014GoingDW, He2015DeepRL, Krizhevsky2012ImageNetCW}. 

\mypara{Step 2: \underline{D}iscriminator \underline{C}onstruction (DC)}
After the first step, the auditor has public artworks, generated artworks, and augmented artworks to train a discriminator. 
For ease of reading, we denote the above three types of artwork as $X_p$, $X_g$, and $X_a$, respectively. 
Recalling the design challenges in ~\autoref{sec:design-challenges}, we leverage a VGG model as the style extractor $\Phi$ and select the outputs of the four evenly spaced layers as the style representations. 
Then, for each artwork, we concatenate the style representations to form the training sample $\Phi(x)$. 
We use 1.0 and $-1.0$ as the target $y$, where $y=1.0$ represents the artwork that originates from the target artist ($y=-1.0$ if it does not). 
Then, the loss function can be formulated as $\left(y-f_{\theta}(\Phi(x))\right)^2$. 
There is a deviation between the original image and the generated image even under the same prompts since the diffusion model has distortion when imitating the artistic style.
This distortion will cause the discriminator to mistakenly judge positive samples as negative. 
Thus, we integrate the distortion in the discriminator's training by measuring the difference between the public artwork and its mimicked version, \ie, $\left(f_{\theta}(\Phi(x_g)) - f_{\theta}(\Phi(x_p)) \right)^2$.
We optimize the weights of $f_{\theta}$ using the following loss function.
\begin{small}
\begin{align}
&\mathcal{L}=\mathcal{L}_{\text{reg}}+\mathcal{L}_{\text{dis}}, \\
&\mathcal{L}_{\text{reg}}=\left(y-f_{\theta}(\Phi(x))\right)^2,  \nonumber \\ 
&\mathcal{L}_{\text{dis}}=\left(f_{\theta}(\Phi(x_g)) - f_{\theta}(\Phi(x_p)) \right)^2,  \nonumber 
\end{align}
\end{small}

\noindent
where $\mathcal{L}_{\text{reg}}$ guides the discriminator in distinguishing between the artworks of the target artist and the artworks of other artists (\ie, $x \in \{X_p, X_a\}$), and the distortion loss $\mathcal{L}_{\text{dis}}$ to calibrate the distortion between the generated artworks and the corresponding original artworks (\ie, $x_g \in X_g, x_p \in X_p$). 

\mypara{Step 3: \underline{A}uditing \underline{P}rocess (AP)}
The auditor conducts the auditing process based on the trained discriminator.
We use the same CLIP interrogator as in Step 1 to create a set of captions.
To encourage the suspicious model to incorporate more features of the target artists in the generated artwork, 
we include the target artists' information in the captions. 
The auditor employs the style extractor to process the generated artworks and obtain their style representations. 
Then, the discriminator predicts the confidence scores based on the style representations. 
Finally, we propose threshold-based and hypothesis-testing-based auditing mechanisms to make the auditing decision.
The auditing mechanisms are detailed in \autoref{sec:the-details-of-audit-process}. 

\subsection{Details of the Auditing Process}
\label{sec:the-details-of-audit-process}
During the auditing process, the discriminator predicts the confidence score based on the multi-granularity style representations from the style extractor.
To improve accuracy, the auditor can utilize several artworks to query the discriminator and aggregate the confidence scores to draw the decision.

A baseline strategy is to compare the average value of the confidence scores with the preset threshold. 
Since the discriminator is a regression model with output ranging from -1 to 1, 
the default threshold is set to 0. 
That is, if the confidence score of an artwork is higher than 0, the auditor will conclude the infringement; otherwise, there is no infringement. 

The other approach involves performing hypothesis testing using the collected confidence scores.
Considering that the confidence scores are continuous, we select the one-sided t-test for hypothesis testing, which is used to determine if the mean of the confidence scores is significantly greater than zero.
\begin{equation}
\begin{aligned}
&H_0: \mu \leq 0, ~\text{The mean value (\(\mu\)) is equal to or less than 0.}   \nonumber \\
&H_1: \mu > 0, ~\text{The mean value (\(\mu\)) is greater than 0.}   \nonumber
\end{aligned}
\end{equation}
For a set of confidence scores $\left\{c_i \mid i=1,2,\dots,n \right\}$, t-test performs the following procedures. 
\begin{itemize}
\item [1)] Calculating $t = \frac{\bar{c}-0}{s / \sqrt{n}}$, where $\bar{c}$ is the average value of the samples, $s$ is the standard deviation of the samples, and $n$ is the number of the samples. 
\item [2)] Setting the critical t-value based on the required confidence level (default 95\%). 
\item [3)] If the calculated t-statistic is greater than the critical t-value, the auditor will reject the null hypothesis, indicating that there is statistically significant evidence that the mean is greater than 0. 
\end{itemize}

\subsection{Discussion}
\label{sec:discussion}
Using multiple layers of CNN to extract image features is indeed a common practice in the computer vision domain.
The subtle difference in this line of methods originates from their different optimized goals and manifests in processing the feature maps derived from the CNN filters.
For instance, Gatys~\etal~\cite{gatys2015neural} aim at the sample-level style transfer task, \ie, each image represents a specific style. They first calculate the correlations between different filter responses (Equation 3 in~\cite{gatys2015neural}) and use the Gram matrix to represent this layer.
Among the layers, they use weighting factors to aggregate the features of different layers (Equation 5 in~\cite{gatys2015neural}).

\sysname is designed for a user-level (or artist-level) style audit, which means that multiple images belonging to the same artist should be identified as one style.
Thus, we design a two-step concatenation, first to concatenate the feature maps in each layer and then to concatenate between different layers, which can better maintain the style extraction of all filters.
To alleviate computational overhead in the discriminator construction of \sysname, we select the maximum and average values of each filter's feature map to participate in the concatenation.
In addition, \sysname does not rely on the knowledge of a single artwork's discrepancy on the suspicious model but learns to discriminate the artists' style. 
Another benefit of ArtistAuditor's style extraction is better transferability across datasets. 
We have validated this effectiveness in ~\autoref{sec:transferability}.

\section{Evaluation}
\label{sec:evaluation}
We first validate the effectiveness of \sysname on three diffusion models, \ie, Stable Diffusion v2.1 (SD-V2)~\cite{stable2-1}, Stable Diffusion XL (SDXL)~\cite{Podell2023SDXLIL}, and Kandinsky~\cite{Razzhigaev2023KandinskyAI} in \autoref{sec:audit-perform}.
We evaluate the transferability of \sysname on different datasets and models in \autoref{sec:transferability}.
Finally, in \autoref{sec:real-world-performance}, we utilize \sysname to audit the text-to-image models fine-tuned on a public platform \href{https://www.scenario.com/}{Scenario}.

\subsection{Experimental Setup}
\label{sec:exp-setup}

\mypara{Target Models} We adopt three text-to-image models, \textit{Stable Diffusion v2.1 (SD-V2)~\cite{stable2-1}}, \textit{Stable Diffusion XL (SDXL)~\cite{Podell2023SDXLIL}}, and  \textit{Kandinsky~\cite{Razzhigaev2023KandinskyAI}}, which are popularly used in previous work~\cite{Shan2023PromptSpecificPA, shan2023glaze, Luo2023StealMA}. 
Due to space limitation, we refer to \autoref{sec:more-details-of-the-selected-models} for more details.

\mypara{Datasets}
We use the WikiArt dataset\footnote{https://www.wikiart.org/} following the prior works~\cite{Wang2024PropertyEI, Cao2023IMPRESSET}, and randomly select fifty artists.
We also build a new dataset, called Artist-30, containing the artworks of thirty artists based on fresh-published datasets~\cite{liao2022artbench} and publicly licensed artworks.
The assessment on Artist-30 highlights \sysname's effectiveness in protecting artworks by lesser-known or emerging artists, who are more susceptible to such attacks than renowned figures like Vincent van Gogh.
\autoref{table:url-source} shows the sources of the collected artworks.
We randomly selected twenty artworks from each artist.

\mypara{Metrics} 
We adopt four metrics, \ie, accuracy, area under the curve (AUC), F1 Score, false positive rate (FPR).
Note that the false positive rate, \ie erroneously labels a model as infringing, can cause reputational harm, financial costs, and strain judicial resources in high-stakes IP litigation.
To mitigate this, \sysname prioritizes minimizing false positives through enhanced accuracy and transparent hypothesis testing.

\mypara{Methods} 
``thold'' is the threshold-based auditing strategy, and ``t-test'' denotes the hypothesis testing-based auditing strategy. 
Both methods share modules except for the decision-making strategy. 
By default, we adopt a median threshold of 0 for \sysname, as it utilizes a regression model to evaluate style representations, assigning a score of 1 for infringement and -1 for non-infringement.
The threshold setting is inherently practical as it does not rely on additional assumptions about the suspicious model.
This improves \sysname's practicality for black-box auditing in real-world applications.
Experimental results show that this threshold works well across datasets and model configurations.

\mypara{Competitors} 
As in~\autoref{sec:difference-with-the-existing-solutions}, the MI methods~\cite{Pang2023BlackboxMI, Wang2024PropertyEI} can be modified to address the data-use auditing. 
Pang~\etal~\cite{Pang2023BlackboxMI} focus on the sample-level inference of the fine-tuning set by the similarity of the original artwork and the generated artwork. 
For each original artwork, Pang~\etal~\cite{Pang2023BlackboxMI} utilize a classifier to predict whether it is a member or not.
We slightly modify this method to align with the requirements of artist-level data-use auditing.
Specifically, after Pang~\etal~\cite{Pang2023BlackboxMI} generate inference results for each artwork by the target artist, we convert their binary predictions into numerical values (1.0 for positive and -1.0 for negative).
Then, these values are used to apply the two auditing mechanisms of \sysname, threshold-based and hypothesis testing-based, to make the final auditing decision.
As both auditing methods yield identical outcomes as in ~\cite{Pang2023BlackboxMI}, we present only one set of results.

We do not consider Wang~\etal~\cite{Wang2024PropertyEI} as a major baseline due to two critical limitations: the lack of open-sourced code (empty GitHub repository) and an AUC of 0.75 for property inference attacks on Stable Diffusion and WikiArt.
Therefore, we opt to present the reproduction settings and experimental results of ~\cite{Wang2024PropertyEI} in~\autoref{sec:comparison-with-wang}.
Regarding watermark-based techniques~\cite{Cui2023DiffusionShieldAW, Cui2023FTShieldAW, Luo2023StealMA}, they require embedding watermarks before the release of artwork, compromising integrity.
Since \sysname performs post hoc auditing without prior modification, these methods are outside our scope.
Thus, we mainly compare the method of ~\cite{Pang2023BlackboxMI} in our evaluations. 

\begin{table}[t]
    \centering
    \caption{The sources of artworks.} 
    \scalebox{0.8}{
    \begin{tabular}{c|l}
        \hline
        \textbf{Artist} & \multicolumn{1}{c}{\textbf{URL Source}} \\
        \hline
        Xia-e & https://huaban.com/boards/58978522 \\
        Fang Li & https://huaban.com/boards/40786095 \\
        Kelek & https://gallerix.asia/storeroom/1725860866 \\
        Norris Joe & https://gallerix.asia/storeroom/1784565901 \\
        Jun Suemi & https://gallerix.asia/storeroom/2000726542 \\
        Geirrod Van Dyke & https://www.artstation.com/geirrodvandyke \\
        Wer & https://www.gracg.com/user/p3133PKMV3r \\
        The remaining 23 artists & https://github.com/liaopeiyuan/artbench \\
        \hline
    \end{tabular}
    \label{table:url-source}
    }
\end{table}

\begin{table*}[!t]
  \centering
  \caption{Overall auditing performance on four evaluation metrics. 
  We report the mean and standard variance of five repeated experiments. 
  ``thold'' is the threshold-based auditing strategy. 
  ``t-test'' denotes the hypothesis testing-based auditing strategy. 
  }
  \vspace{-0.3cm}
  \scalebox{0.72}{
  \begin{tabular}{c|c|ccc|ccc|ccc}
  \hline
  \multirow{3}{*}{\textbf{Dataset}}          & \textbf{Model}                            & \multicolumn{3}{c|}{\textbf{SD-V2}}                         & \multicolumn{3}{c|}{\textbf{SDXL}}                                       & \multicolumn{3}{c}{\textbf{Kandinsky}}                                                                                          \\ \cline{2-11} 
                                             & \diagbox[dir=NW]{\textbf{Metric}}{\textbf{Method}} & \textbf{Pang et al.~\cite{Pang2023BlackboxMI}} & \textbf{thold}  & \textbf{t-test} & \textbf{Pang et al.~\cite{Pang2023BlackboxMI}} & \textbf{thold}     & \textbf{t-test} & \textbf{Pang et al.~\cite{Pang2023BlackboxMI}} & \textbf{thold}  & \textbf{t-test} \\ \hline
  \multirow{4}{*}{\textbf{WikiArt}}          & Accuracy                                           & 0.733$\pm$0.019                                                            & \textbf{0.908$\pm$0.020}          & 0.896$\pm$0.015             & 0.813$\pm$0.009                    & 0.852$\pm$0.010          & \textbf{0.868$\pm$0.010} & 0.793$\pm$0.025                    & \textbf{0.892$\pm$0.020}       & 0.852$\pm$0.010 \\
                                             & AUC                                                & 0.838$\pm$0.022                                                            & \textbf{0.967$\pm$0.007}          & /                           & 0.885$\pm$0.013                    &\textbf{ 0.937$\pm$0.003} & /                        & \textbf{1.000$\pm$0.000}           & 0.973$\pm$0.004                & /               \\
                                             & F1 Score                                           & 0.661$\pm$0.027                                                            & \textbf{0.915$\pm$0.018}          & 0.895$\pm$0.015             & 0.803$\pm$0.028                    & 0.866$\pm$0.008          & \textbf{0.875$\pm$0.008} & 0.802$\pm$0.006                    & \textbf{0.888$\pm$0.020}       & 0.826$\pm$0.014 \\
                                             & FPR                                                & 0.107$\pm$0.019                                                            & 0.176$\pm$0.041                   & \textbf{0.096$\pm$0.032}    & 0.293$\pm$0.050                    & 0.256$\pm$0.020          & \textbf{0.184$\pm$0.020} & 0.493$\pm$0.019                    & 0.072$\pm$0.030                & \textbf{0.000$\pm$0.000} \\ \hline
  \multirow{4}{*}{\textbf{Artist-30}}        & Accuracy                                           & 0.767$\pm$0.027                                                            & \textbf{0.953$\pm$0.045}          & 0.880$\pm$0.045             & 0.800$\pm$0.027                    & \textbf{0.947$\pm$0.016} & 0.867$\pm$0.021          & 0.922$\pm$0.016                    & 0.933$\pm$0.021                & \textbf{0.973$\pm$0.025} \\
                                             & AUC                                                & 0.986$\pm$0.004                                                            & \textbf{0.992$\pm$0.009}          & /                           & 0.923$\pm$0.030                    & \textbf{1.000$\pm$0.000} & /                        & \textbf{1.000$\pm$0.000}           & 0.998$\pm$0.004                & /               \\
                                             & F1 Score                                           & 0.694$\pm$0.046                                                            & \textbf{0.951$\pm$0.049}          & 0.864$\pm$0.054             & 0.749$\pm$0.043                    & \textbf{0.943$\pm$0.018} & 0.845$\pm$0.028          & 0.909$\pm$0.000                    & 0.938$\pm$0.019                & \textbf{0.975$\pm$0.023} \\
                                             & FPR                                                & \textbf{0.000$\pm$0.000}                                                   & 0.027$\pm$0.033                   & 0.013$\pm$0.027             & 0.000$\pm$0.000                    & \textbf{0.000$\pm$0.000} & 0.000$\pm$0.000          & 0.200$\pm$0.000                    & 0.133$\pm$0.042                & \textbf{0.053$\pm$0.050} \\ \hline
  \end{tabular}}
  \label{table:audit-performance}
\end{table*}

\mypara{Default Experimental Settings}
In the evaluation, we use the following experimental settings as the default if there is no additional statement. 
We randomly split the artists into two groups and utilized the artworks created by the first group to fine-tune the diffusion model. 
For ease of reading, we note the first group of artworks as $D^{+}$ and the second group of artworks as $D^{-}$. 
We use the CLIP interrogator to generate a description for each artwork and include the artist's name in the caption, following the previous work~\cite{shan2023glaze}. 
We fine-tune the target model using the dataset $D^{+}$. 
During the training of each artist's discriminator, we use the original artworks of each artist as positive samples and further divide them into training samples and validation samples at a ratio of 8:2. 
For negative samples, we randomly select from the other artists' artworks while keeping a positive-to-negative ratio of 1:1. 

\begin{itemize}
    \item \textit{The Settings of Fine-tuning}: Following the previous work~\cite{Cao2023IMPRESSET}, we use the corresponding fine-tuning scripts released with the models~\cite{von-platen-etal-2022-diffusers}. %
    More specifically, SD-V2 is fine-tuned for 100 epochs on the dataset $D^{+}$ using the AdamW optimizer with a learning rate of $5\times10^{-6}$. 
    SDXL is fine-tuned for 100 epochs on the dataset $D^{+}$ using the AdamW optimizer with a learning rate of $1\times10^{-4}$.
    As for Kandinsky, both the prior and decoder are fine-tuned for 100 epochs on the dataset $D^{+}$ using the AdamW optimizer with a learning rate of $1\times10^{-4}$.
    \item \textit{The Settings of Discriminator}: 
    We optimize the discriminator by Adam optimizer with a learning rate of $5\times10^{-5}$.
    The entire training takes 100 epochs, and we utilize an early stopping method with a patience of 10.  
\end{itemize}

\subsection{Overall Auditing Performance}
\label{sec:audit-perform}
We assess the auditing effectiveness of \sysname and its competitor~\cite{Pang2023BlackboxMI} for SD-V2, SDXL, and Kandinsky.

\mypara{Setup}
We collect 20 prompts for each artist and query the target model to obtain 20 generated images. 
Then, the auditor puts the images into the style extractor, converts them into style representations, and gets the corresponding confidence scores based on the discriminator.
Finally, we combine the auditing results of 20 artists to calculate the accuracy, AUC, F1 score, and FPR values.
The experimental results are in \autoref{table:audit-performance}, where the values of mean and standard variation are calculated by repeating the experiment 5 times with five random seeds $\{1, 2, 3, 4, 5\}$.

\mypara{Observations}
We have the following observations from \autoref{table:audit-performance}. 
1) \sysname archives consistent high auditing performance. 
The accuracy values are higher than 0.852 for all models. 
These results indicate that \sysname is highly effective in identifying unauthorized use of artists' artworks for different diffusion models.
In addition, the AUC values are almost perfect for all models, \ie, more than 0.937.
2) The AUC values of \sysname fluctuate in different combinations of models and datasets.
\sysname achieves a remarkable AUC on Artist-30 (AUC = 1), while \sysname obtains a lower AUC of 0.937 on WikiArt.
We speculate that the reason is that SDXL's pre-training process uses a part of the internal dataset, which may overlap with the artworks in WikiArt. 
When using the same fine-tuning dataset, the AUC values of \sysname vary on different models, such as SDXL and Kandinsky.
Compared with SD-V2 and SDXL, Kandinsky switches to CLIP-ViT-G as the image encoder, significantly increasing the model's capability to generate more aesthetic pictures. 
3) The FPR values of  ``t-test'' usually lower than those of ``thold''.
The selection of the threshold is an empirical process, and the average confidence score is easily misled by the outlier. 
Compared to ``thold'', ``t-test'' calculates the statistic $t$, where the number and variance of confidence scores are also considered in the hypothesis testing. 
4) \sysname is superior to the competitor in most experimental settings.
The accuracy values of \sysname are generally higher than those of ~\cite{Pang2023BlackboxMI} with a lower FRP.
The reason is that ~\cite{Pang2023BlackboxMI} aims at the features of the individual samples in the fine-tuning set, ignoring the commonality in style between the artworks of the same artist.
Pang~\etal~\cite{Pang2023BlackboxMI} cannot deal with the situation where the artworks used to fine-tune the suspicious model are inconsistent with the artworks used for auditing.
This can be further corroborated by the results on the transferability of the dataset of \autoref{sec:transferability}.

\subsection{Transferability of \sysname}
\label{sec:transferability}
The auditor is unaware of the selected artworks or the image captioning model used to fine-tune the suspicious model.
Therefore, this section aims to assess the transferability of \sysname.
We begin by evaluating the dataset transferability when the artworks used for auditing differ from those used to fine-tune the suspicious model.
Next, we assess model transferability when the auditor's image captioning model differs from that of the suspicious model.

\mypara{Dataset Transferability}
We consider two scenarios, \ie, the partial overlap and the disjoint cases.
In the partial overlap scenario, the artworks used by the suspicious model overlap half with the artworks used by the auditor.
In the disjoint scenario, the auditor has a set of artworks by the target artist.
These artworks are different from the artworks used in the fine-tuning of the suspicious model.
For each experimental setting, we perform five replicate experiments with random seeds set to \{1, 2, 3, 4, 5\}.
We then report the mean and variance of the results.

\autoref{table:dataset-transferability} shows the effectiveness of \sysname in auditing piracy of artistic style across different degrees of overlap in the dataset.
1) When the artworks partially overlap, the performance of \sysname slightly decreases.
\sysname still remains effective with AUC $>0.964$ and FPR $<0.133$.
For example, for the SDXL model, \sysname achieves an auditing accuracy of up to 0.920, which is only 0.027 lower than that of the complete overlap scenario. 
This indicates the internal consistency of the artist's work style, which can be extracted by the style extractor and used as an auditing basis for whether infringement of the artwork occurs.

2) The most significant performance drop is observed in the disjoint scenario, particularly in accuracy and F1 scores.
Compared to~\cite{Pang2023BlackboxMI}, \sysname can still detect the mimicked artworks.
Especially on the Kandinsky model, \sysname's auditing accuracy only drops by 0.026 compared to the complete overlap scenario.
The comparison demonstrates that \sysname does not rely on the overfitting of individual artwork but rather learns to discriminate based on the internal commonality of the artist's style. 

\begin{table}[!t]
\centering
\caption{Dataset Transferability of \sysname. 
``Partially'' and ``Disjoint'' refer to the dataset's partial overlap and disjoint scenarios. 
\autoref{table:full-dataset-transferability} shows more details.  
} 
\vspace{-0.2cm}
\scalebox{0.8}{
\begin{tabular}{c|c|ccc|ccc}
\hline
\multirow{2}{*}{\textbf{Model}}     & \textbf{Setting} & \multicolumn{3}{c|}{\textbf{Partially}}                               & \multicolumn{3}{c}{\textbf{Disjoint}}                                 \\ \cline{2-8} 
                                    & \textbf{Metric}  & \textbf{\cite{Pang2023BlackboxMI}} & \textbf{thold} & \textbf{t-test} & \textbf{\cite{Pang2023BlackboxMI}} & \textbf{thold} & \textbf{t-test} \\ \hline
\multirow{4}{*}{\textbf{SD-V2}}     & Accuracy         & 0.789                              & \textbf{0.800}          & 0.760           & 0.556                              & \textbf{0.727}          & 0.687           \\
                                    & AUC              & \textbf{0.991}                              & 0.964          & /               & 0.699                              & \textbf{0.956}         & /               \\
                                    & F1 Score         & 0.745                              & \textbf{0.754}          & 0.683           & 0.281                              & \textbf{0.623}          & 0.543           \\
                                    & FPR              & 0.000                              & 0.000          & 0.000           & 0.000                              & 0.000          & 0.000           \\ \hline
\multirow{4}{*}{\textbf{SDXL}}      & Accuracy         & 0.689                              & \textbf{0.920}          & 0.873           & 0.511                              & \textbf{0.727}          & 0.633           \\
                                    & AUC              & 0.921                              & \textbf{1.000}          & /               & 0.872                              & \textbf{0.980}          & /               \\
                                    & F1 Score         & 0.576                              & \textbf{0.912}          & 0.855           & 0.148                              & \textbf{0.622}          & 0.419           \\
                                    & FPR              & 0.000                              & 0.000          & 0.000           & 0.000                              & 0.000          & 0.000           \\ \hline
\multirow{4}{*}{\textbf{Kandinsky}} & Accuracy         & 0.933                              & 0.933          & \textbf{0.967}           & 0.711                              & \textbf{0.907}          & 0.853           \\
                                    & AUC              & 0.936                              & \textbf{0.996}          & /               & 0.744                              & \textbf{0.982}          & /               \\
                                    & F1 Score         & 0.923                              & 0.938          & \textbf{0.967}           & 0.667                              & \textbf{0.896}          & 0.826           \\
                                    & FPR              & 0.187                              & 0.133          & \textbf{0.053}           & 0.190                              & 0.013          & \textbf{0.000}           \\ \hline
\end{tabular}
}
\label{table:dataset-transferability}
\vspace{-0.3cm}
\end{table}

\begin{figure}[!t]
\centering
\includegraphics[width=0.95\hsize]{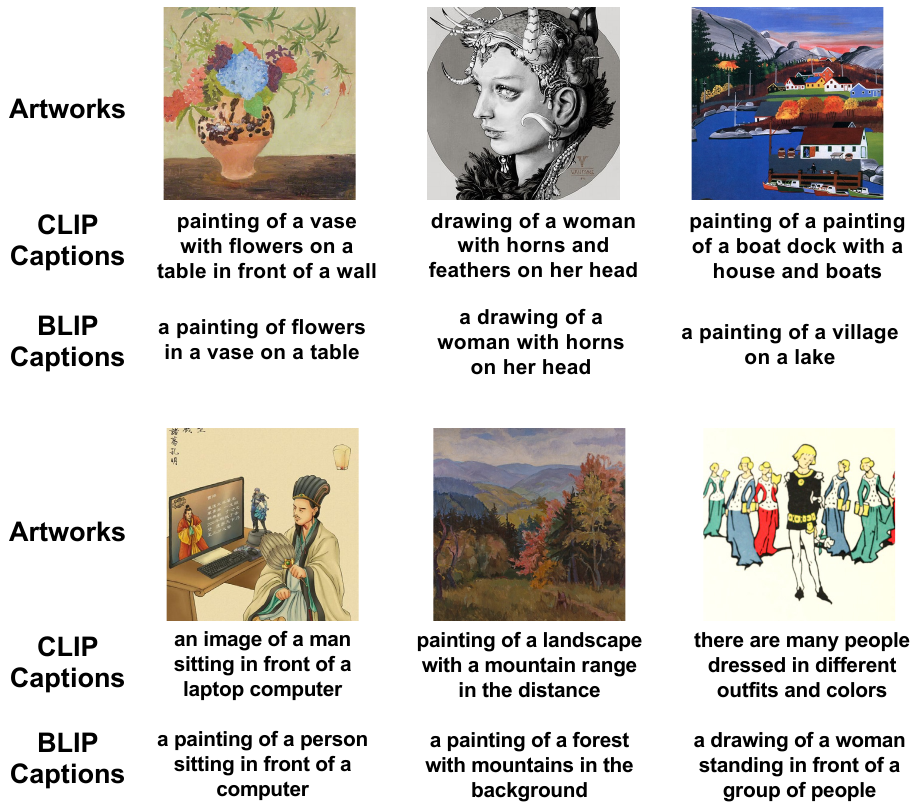}
\caption{CLIP and BLIP generate captions for the same set of artworks, respectively.
} 
\vspace{-0.3cm}
\label{fig:image-prompt}
\end{figure}

\begin{figure}[!t]
    \centering
    \includegraphics[width=\linewidth]{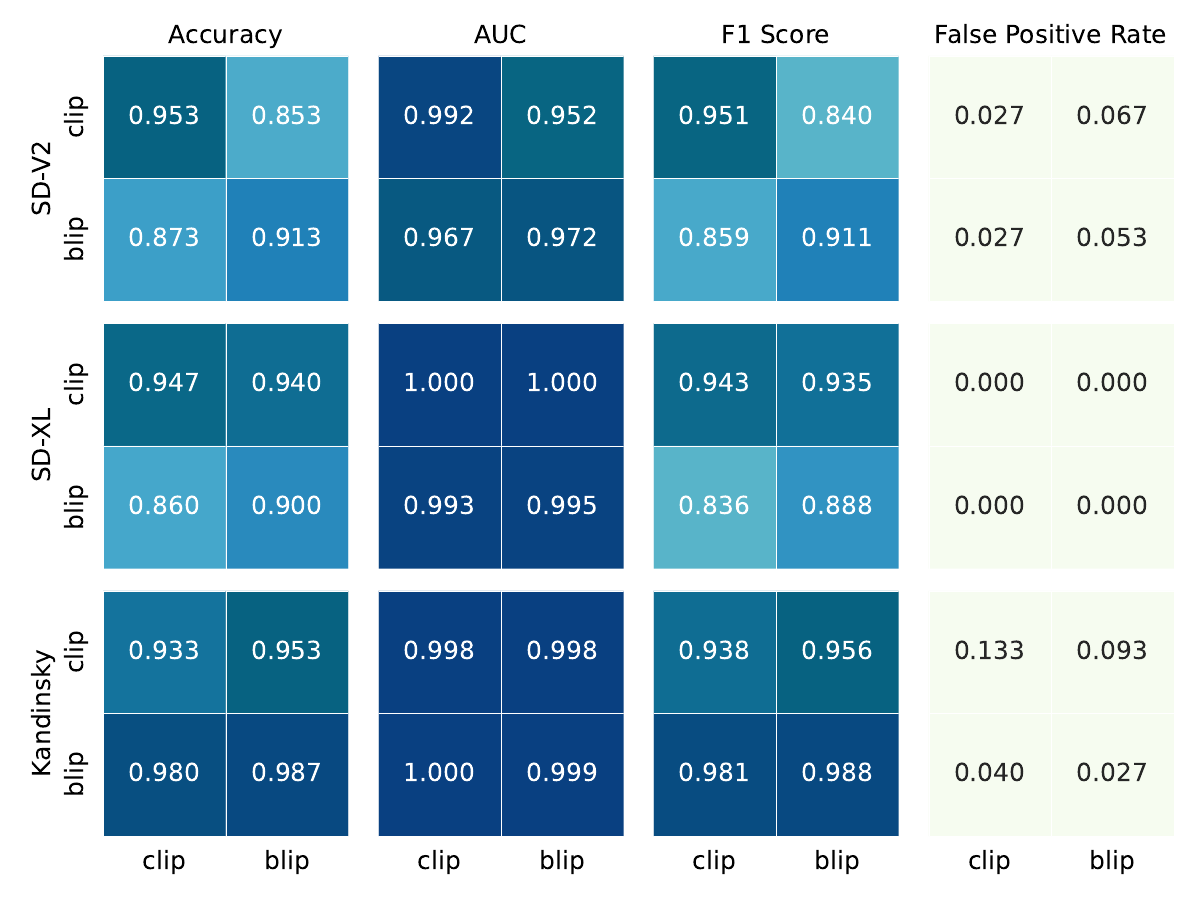} 
    \caption{Model Transferability of \sysname. 
    The x-axis is the image captioning model used in suspicious models. The y-axis is the image captioning model used by the auditor.
    \autoref{table:full-model-transferability} shows more details. 
    }
    \vspace{-0.4cm}
    \label{fig:model-transfer}
\end{figure}

\mypara{Model Transferability}
The suspicious model may apply a different captioning model from that of the auditor to generate prompts.
\autoref{fig:image-prompt} compares the captions generated by two different image captioning models, \ie, CLIP~\cite{Radford2021LearningTV} and BLIP~\cite{Li2022BLIPBL}. 
In every experimental setup, we perform five duplicate trials using random seeds \{1, 2, 3, 4, 5\} and provide the mean and variance of the outputs.

\autoref{fig:model-transfer} shows the model transferability of \sysname.
1) When the same image captioning model is used by both the suspicious model and the auditor, \sysname achieves high auditing performance.
For example, \sysname performs an auditing accuracy of 0.947 on the SDXL model with an FPR equal to 0.
Kandinsky has a higher FPR (0.133) but maintains reasonable accuracy (0.933) and F1 scores (0.938).
2) The results show a slight decrease in auditing performance when different image captioning models are used. 
This is particularly evident in the SD-V2 model, where the accuracy value drops from 0.953 to 0.853, and the F1 score drops from 0.951 to 0.840.
However, the AUC values remain high, indicating strong discriminative power despite the variation in prompt generation.
On one hand, when the artwork's content is fixed, the distribution of suitable captions is limited.
On the other hand, \sysname mainly grasps the stylistic characteristics of the artist rather than fitting specific artwork, making it robust to the caption's changes.

\section{Discussion}
\mypara{Highlights of \sysname}
1) \sysname is the first data usage auditing method for the diffusion model without the requirement of the model's retraining or modification to original artworks. 
2) By comprehensively evaluating \sysname in different experimental settings, such as dataset transferability, model transferability, data augmentation, and  distortion calibration, 
we conclude some useful observations for adopting \sysname. 
3) We apply \sysname to audit fine-tuned models on an online platform. 
The auditing decisions are all correct, demonstrating that \sysname is an effective and efficient strategy for practical use. 

\mypara{Limitations and Future Work}
We discuss the limitations of \sysname and promising directions for further improvements. 
1) From \autoref{sec:audit-perform}, the accuracy of \sysname decreases when more artists' works are involved in the fine-tuning process. 
Thus, it is interesting to enhance \sysname by mining more personalized features from the artists' works. 
2) Adversarial perturbation may decrease the auditing accuracy of \sysname, such as differential privacy~\cite{wang2024dpadapter, wei2023dpmlbench, yuan2023privgraph} and image compression~\cite{Dziugaite2016ASO}. 
Preliminary experiments on the SD-V2 model show that AUC drops from 0.992 to 0.921 when the generated images undergo JPEG compression (quality level: 20).
With adversarial training, the score improves to 0.980.
Thus, adversarial training is a promising mitigation approach. 

\section{Conclusion}
In this work, we propose a novel artwork auditing method for text-to-image models based on the insight that the multi-granularity latent representations of a CNN model can serve as the intrinsic fingerprint of an artist.
Through extensive experiments, we show that \sysname is an effective and efficient solution for protecting the intellectual property of artworks.
The experimental results on six combinations of models and datasets show that \sysname can achieve high AUC values (>~0.937).
The auditing process can be performed on a consumer-grade GPU.
We conclude several important observations for adopting \sysname in practice by evaluating the dataset transferability, the captioning model transferability, the impact of data augmentation, and the impact of distortion calibration.
Finally, we utilize the online commercial platform~\href{https://www.scenario.com/}{Scenario} to examine the practicality of \sysname, and show that \sysname behaves excellently on real-world auditing.

\begin{acks}
We would like to thank the anonymous reviewers for their constructive comments. 
This work was partly supported by the National Key Research and Development Program of China under No. 2022YFB3102100, the NSFC under Grants No. 62293511, 62402379, U244120033, U24A20336 and 62402425, the Zhejiang Province Science Foundation under Grants LD24F020002, and the Key Research and Development Program of Zhejiang Province (Grant Number: 2024C01SA160196).
Zhikun Zhang was supported in part by the NSFC under Grants No. 62441618 and Zhejiang University Education Foundation Qizhen Scholar Foundation.
Min Chen was partly supported by the project CiCS of the research programme Gravitation which is (partly) financed by the Dutch Research Council (NWO) under Grant No. 024.006.037.
\end{acks}

\newpage
\balance
\bibliographystyle{abbrv}
\bibliography{refs}

\appendix

\section{Difference with the Existing Solutions}
\label{sec:difference-with-the-existing-solutions}
Recent works~\cite{Wang2024PropertyEI, Kandpal2023UserIA, Pang2023BlackboxMI, Duan2023AreDM} study MI methods against diffusion models.  
These methods can be adapted to solve the data-use auditing task. 
Among these, the strategies~\cite{Pang2023BlackboxMI, Wang2024PropertyEI}, which are designed for black-box settings, are notable for their state-of-the-art performance. 
However, \sysname differs from these strategies in several essential aspects.
In~\autoref{table:overall-comparison}, we provide an overall comparison between the existing works and \sysname.
It is worth noting that these differences are mainly since they are optimized for different inference objectives. 
That is, Pang~\etal~\cite{Pang2023BlackboxMI} is for individual samples in the fine-tuning dataset. 
Wang~\etal~\cite{Wang2024PropertyEI} works for the concrete property among the training samples. 
\sysname is optimized for the abstract property, \ie, artist's style. 

\begin{itemize}[leftmargin=*]
    \item \mypara{Feature Extraction} 
    The artwork's style is typically defined by a complex blend of elements, including low-level brushstrokes and high-level painterly motifs. 
    Compared to \cite{Pang2023BlackboxMI}, \sysname makes the final judgment by concatenating the features of different layers, thus better portraying the artist's artistic style. 
    \item \mypara{Similarity Measurement} 
    Wang~\etal~\cite{Wang2024PropertyEI} derives inference results by calculating the cosine similarity between anchor images and generated images, which is appropriate for dealing with concrete property in an image. 
    However, artistic style is a more abstract concept. 
    For instance, despite having completely different subjects, ``Wheatfield with Crows'' and ``The Starry Night'' both belong to the same painter, Van Gogh. 
    Thus, we leverage an MLP model to portray the similarity of styles and derive auditing results based on the confidence scores of multiple artworks. 
    \item \mypara{Distortion Calibration} 
    Due to the limitations of the model capability and the influence of other artists' artworks in the pre-training dataset, the generated artworks inevitably suffer artistic distortions. 
    Compared to \cite{Wang2024PropertyEI, Pang2023BlackboxMI}, \sysname considers this distortion, reducing the omission of potential infringements. 
\end{itemize}

\begin{table}[!t]
    \centering
    \caption{Overview of the existing methods for data copyright protection. 
    `Tech.' refers to the core technology used by the method. 
    `DA' (Data Access) refers to whether the method needs access to the image or both the image and the corresponding prompt. 
    `DF' (Data Fidelity) stands for whether the method maintains data fidelity or not. 
    `TD' (Training Data) refers to whether the method needs access to the training data of the suspicious model. 
    `SM' (Shadow Model) refers to whether the method requires training shadow models.} 
    \scalebox{0.8}{
    \begin{tabular}{ccccccc}
    \hline
    \textbf{Method}                                        & \textbf{Goal}                                                                & \textbf{Tech.}                                                                      & \textbf{DA} & \textbf{DF}  & \textbf{TD}  & \textbf{SM}  \\ \hline
    \cite{shan2023glaze}                                   & \multirow{2}{*}{\begin{tabular}[c]{@{}c@{}}Preventing\\ misuse\end{tabular}} & \multirow{2}{*}{\begin{tabular}[c]{@{}c@{}}Adversarial\\ perturbation\end{tabular}} & Image       & $\times$     & $\times$     & $\times$ \\
    \cite{Van_Le_2023_ICCV}                                &                                                                              &                                                                                     & Image       & $\times$     & $\times$     & $\times$ \\ \cline{2-7} 
    \cite{Ma2023GenerativeWA}                              & \multirow{7}{*}{\begin{tabular}[c]{@{}c@{}}Detecting\\ misuse\end{tabular}}  & \multirow{3}{*}{\begin{tabular}[c]{@{}c@{}}Backdoor-based\\ watermark\end{tabular}} & Both        & $\times$     & $\times$     & $\checkmark$     \\
    \cite{Cui2023DiffusionShieldAW}                        &                                                                              &                                                                                     & Image       & $\times$     & $\times$     & $\times$ \\
    \cite{wang2023diagnosis}                               &                                                                              &                                                                                     & Image       & $\times$     & $\times$     & $\times$ \\ \cline{3-7} 
    \cite{Chen2019GANLeaksAT}                              &                                                                              & \multirow{4}{*}{\begin{tabular}[c]{@{}c@{}}Membership \\ inference\end{tabular}}    & Both        & $\checkmark$ & $\checkmark$     & $\times$ \\
    \cite{Wang2024PropertyEI}                              &                                                                              &                                                                                     & Both        & $\checkmark$ & $\checkmark$     & $\times$ \\
    \cite{Pang2023BlackboxMI}                              &                                                                              &                                                                                     & Image       & $\checkmark$ & $\times$     & $\checkmark$     \\ \cline{4-7} 
    Ours                                                   &                                                                              &                                                                                     & Image       & $\checkmark$ & $\times$     & $\times$ \\ \hline
    \end{tabular}}
    \label{table:overall-comparison}
\end{table}

\section{More Details of the Text-to-Image Models}
\label{sec:more-details-of-the-selected-models}
We provide more details of the three text-to-image models used in~\autoref{sec:evaluation}.
\begin{itemize}
    \item \textit{Stable Diffusion v2.1 (SD-V2)~\cite{stable2-1}}: SD-V2 is a high-performing and open-source model, trained on 11.5 million images from LAION~\cite{schuhmann2022laion}. 
    It achieves state-of-the-art performance on several benchmarks \cite{Rombach_2022_CVPR}. 
    \item \textit{Stable Diffusion XL (SDXL)~\cite{Podell2023SDXLIL}}: SDXL represents the latest advancement in diffusion model, significantly outpacing its predecessor, SD-V2, across multiple performance benchmarks. 
    This model boasts a substantial increase in complexity, containing over 2.6 billion parameters, a stark contrast to the 865 million parameters of SD-V2. 
    Compared to SD-V2, SDXL introduces a refiner structure to enhance the quality of image generation.
    \item \textit{Kandinsky~\cite{Razzhigaev2023KandinskyAI}}: Kandinsky is a novel text-to-image synthesis architecture that combines image-prior models with latent diffusion techniques. 
    An image prior model, which is separately trained, maps text embeddings to image embeddings using the CLIP model. 
    Kandinsky also features a modified MoVQ implementation serving as the image autoencoder component. 
\end{itemize}

\section{Data Augmentation}
\label{sec:data-augmentation}
This section elaborates on the data augmentation strategies used in \autoref{sec:workflow-of-auditor}. 

\begin{itemize}
    \item \mypara{Random Cropping} It involves selecting a random portion of the image and using only that cropped part for training, which helps the model focus on different parts of the image and learn more comprehensive features. 
    \item \mypara{Random Horizontal Flipping} This augmentation technique flips images horizontally at random. This is particularly useful for teaching the model that the orientation of objects can vary, and it should still be able to recognize the object regardless of its mirrored position. 
    \item \mypara{Random Cutouts} It involves randomly removing squares or rectangles of various sizes from an image during training. This forces the model to focus on less information and learn to make predictions based on partial views of objects. It is beneficial for enhancing the model's ability to focus on the essential features of the image without overfitting to specific details. 
    \item \mypara{Gaussian noise} It injects noise that follows a Gaussian distribution into image pixels. This technique helps the model become more robust to variations in pixel values and can improve its ability to generalize well on new, unseen data. 
    \item \mypara{Impulse noise} Impulse noise, also known as salt-and-pepper noise, randomly alters the pixel values in images, turning some pixels completely white or black. Training with impulse noise can help the model learn to ignore significant but irrelevant local variations in the image data. 
    \item \mypara{Color jittering} It encompasses adjustments to brightness, saturation, contrast, and hue of the image randomly, which is beneficial for preparing the model to handle images under various lighting conditions and color settings. 
\end{itemize}

\section{Ablation Study}
\label{sec:ablation-study}

\mypara{Impact of Data Augmentation}
Recalling \autoref{sec:workflow-of-auditor}, the data augmentation aims to expand the number of artworks for training discriminators. 
We compare the performance of \sysname with and without data augmentation. 

The results in columns ``w/o DA'' and ``Baseline'' of \autoref{table:DA-and-DC-impact} show that data augmentation significantly enhances auditing performance. 
For instance, the accuracy of \sysname increases from 0.633 to 0.947 in the SDXL model, and from 0.647 to 0.933 in the Kandinsky model. 
Data augmentation significantly increases the number and diversity of artworks, preventing the discriminator from overfitting to style-irrelevant features.

\begin{table}[!t]
  \centering
  \caption{Impact of data augmentation and distortion calibration. 
  ``w/o DA'' shows the auditing performance without data augmentation. 
  ``w/o DC'' shows the auditing performance without distortion calibration.
  \autoref{table:full-DA-and-DC-impact} shows more details. 
  } 
  \scalebox{0.76}{
  \begin{tabular}{c|c|cc|cc|cc}
      \hline
      \multirow{2}{*}{\textbf{Model}} & \textbf{Setting} & \multicolumn{2}{c|}{\textbf{w/o DA}}  & \multicolumn{2}{c|}{\textbf{w/o DC}} & \multicolumn{2}{c}{\textbf{Baseline}} \\
      \cline{2-8}
      ~ & \textbf{Metric} & \textbf{thold} & \textbf{t-test} & \textbf{thold} & \textbf{t-test} & \textbf{thold} & \textbf{t-test}\\
      \hline
      \multirow{4}{*}{\textbf{SD-V2}}      & Accuracy            & 0.953   & 0.853   & 0.927 & 0.867 & 0.953 & 0.880 \\
                                  & AUC                 & 0.994   & /       & 0.995 & /     & 0.992 & / \\
                                  & F1 Score            & 0.951   & 0.825   & 0.920 & 0.845 & 0.951 & 0.864 \\
                                  & FPR & 0.013   & 0.000   & 0.000 & 0.000 & 0.027 & 0.013 \\
      \hline                                                        
      \multirow{4}{*}{\textbf{SDXL}}       & Accuracy            & 0.953   & 0.893   & 0.633 & 0.620 & 0.947 & 0.867 \\
                                  & AUC                 & 0.997   & /       & 0.874 & /     & 1.000 & / \\ 
                                  & F1 Score            & 0.951   & 0.879   & 0.411 & 0.372 & 0.943 & 0.845 \\
                                  & FPR & 0.000   & 0.000   & 0.000 & 0.000 & 0.000 & 0.000 \\
      \hline                                                        
      \multirow{4}{*}{\textbf{Kandinsky}}  & Accuracy            & 0.880   & 0.913   & 0.647 & 0.620 & 0.933 & 0.973 \\
                                  & AUC                 & 0.977   & /       & 0.850 & /     & 0.998 & / \\
                                  & F1 Score            & 0.893   & 0.920   & 0.460 & 0.382 & 0.938 & 0.975 \\
                                  & FPR & 0.240   & 0.173   & 0.013 & 0.000 & 0.133 & 0.053 \\
      \hline
  \end{tabular}}
  \label{table:DA-and-DC-impact}
\end{table}

\mypara{Impact of Distortion Calibration}
In \autoref{sec:workflow-of-auditor}, we try to calibrate the style distortion between the artworks generated by the suspicious model and the original artworks used in its training process. 
The calibration dataset comprises artworks from two sources: public artworks and generated artworks. 
We evaluate the impact of distortion calibration. 

The results in columns ``w/o DC'' and ``Baseline'' of \autoref{table:DA-and-DC-impact} show that the distortion calibration generally improves accuracy for both auditing strategies. 
For example, the auditing accuracy of \sysname on Kandinsky increases from 0.880 to 0.933, while the FPR decreases from 0.240 to 0.133. 
With the help of distortion calibration, the discriminator can effectively learn the subtle differences between the style of original artworks and the style of model-generate artworks. 
This makes \sysname more robust in detecting unauthorized usage, ensuring better protection of IP. 

\section{Comparison with Wang~\etal~\cite{Wang2024PropertyEI}}
\label{sec:comparison-with-wang}
The property existence
inference method~\cite{Wang2024PropertyEI} has three stages: property extractor training, similarity computation, and threshold selection.
Thus, we introduce the reproduction of ~\cite{Wang2024PropertyEI} following its attack procedure.

In the training of property extractors, Wang~\etal~\cite{Wang2024PropertyEI} employ a deep learning model that utilizes the triplet loss function~\cite{schroff2015facenet}.
Specifically, the model is trained to reduce the cosine distance between the base and the positive embeddings, and to increase it between the base and the negative embeddings.
To align with \sysname, we instantiate the property extractor using the VGG model.
Then, we construct the base, the positive, and the negative embeddings based on the target artist's artworks and the public artworks in ~\autoref{fig:framework}.
Concretely, we randomly pair each two artworks from the target artist to form the base embedding and the positive embedding.
The negative embedding can be obtained from the artworks of other artists.

In the similarity computation stage, Wang~\etal~\cite{Wang2024PropertyEI} calculate a score for the target property by measuring the similarities between the embeddings of the target artworks and those produced by the suspicious model.
Following Algorithm 1 of~\cite{Wang2024PropertyEI}, we use the target artist's works as $D_A$, the suspicious model's generated artworks as $D_{gen}$, and the public artworks as $D_{out}$.
For the hyperparameters $\alpha$ and $K$, we utilize the grid search method to select the best performing setting, where $\alpha = 0.16$ and $K = 3$.

In threshold selection, we adopt the guideline in ~\cite{Wang2024PropertyEI}, \ie, training shadow models to determine a threshold for the final decision.

The remaining experimental setups are consistent with those in~\autoref{sec:audit-perform}.
\autoref{tab:comparison-wang-overall} provides the mean and the standard variance on the four metric.
The results demonstrate that \sysname can achieve better auditing performance than the property inference method in~\cite{Wang2024PropertyEI}.

\begin{table}[!t]
\centering
\caption{Overall auditing performance of Wang~\etal~\cite{Wang2024PropertyEI} on four evaluation metrics.} 
\scalebox{0.76}{
\begin{tabular}{c|c|c|c}
\hline
{\textbf{Model}}     & \diagbox[dir=NW]{\textbf{Metric}}{\textbf{Dataset}}    & {\textbf{WikiArt}}               & {\textbf{Artist-30}}                       \\ 
\hline
\multirow{4}{*}{\textbf{SD-V2}}     & Accuracy         & 0.513$\pm$0.025     & 0.489$\pm$0.016 \\
                                    & AUC              & 0.488$\pm$0.022     & 0.453$\pm$0.035          \\
                                    & F1 Score         & 0.138$\pm$0.046     & 0.109$\pm$0.086 \\
                                    & FPR              & 0.053$\pm$0.075     & 0.089$\pm$0.031          \\ 
\hline
\multirow{4}{*}{\textbf{SDXL}}      & Accuracy         & 0.487$\pm$0.041     & 0.489$\pm$0.042 \\
                                    & AUC              & 0.470$\pm$0.009     & 0.450$\pm$0.058 \\
                                    & F1 Score         & 0.094$\pm$0.067     & 0.116$\pm$0.008 \\
                                    & FPR              & 0.080$\pm$0.057     & 0.089$\pm$0.083          \\ 
\hline
\multirow{4}{*}{\textbf{Kandinsky}} & Accuracy         & 0.520$\pm$0.033     & 0.511$\pm$0.042          \\
                                    & AUC              & 0.535$\pm$0.055     & 0.613$\pm$0.077 \\
                                    & F1 Score         & 0.174$\pm$0.100     & 0.105$\pm$0.149         \\
                                    & FPR              & 0.067$\pm$0.050     & 0.044$\pm$0.031          \\ 
\hline
\end{tabular}
}
\label{tab:comparison-wang-overall}
\vspace{-0.3cm}
\end{table}

\section{Real-World Performance}
\label{sec:real-world-performance}

We demonstrate the effectiveness of \sysname in real-world applications by an online model fine-tuning platform \href{https://www.scenario.com/}{Scenario}. 
After the user uploads a set of artworks, the platform fine-tunes a model to mimic the artistic style and returns an API for the user to generate mimicked artworks. 

\mypara{Setup}
Recalling \autoref{sec:transferability}, the auditor is not aware of the specific artworks used to fine-tune the suspicious model.
Thus, aligning with \autoref{table:dataset-transferability}, we provide the auditing performance in complete overlap, partial overlap, and disjoint cases. 
Due to the limited number of images for single fine-tuning on \href{https://www.scenario.com/}{Scenario}, we randomly pick 10 artworks from each artist and upload them to fine-tune the model. 
We perform auditing for three different artists separately.

\begin{figure*}[htbp]
\centering
\includegraphics[width=\hsize]{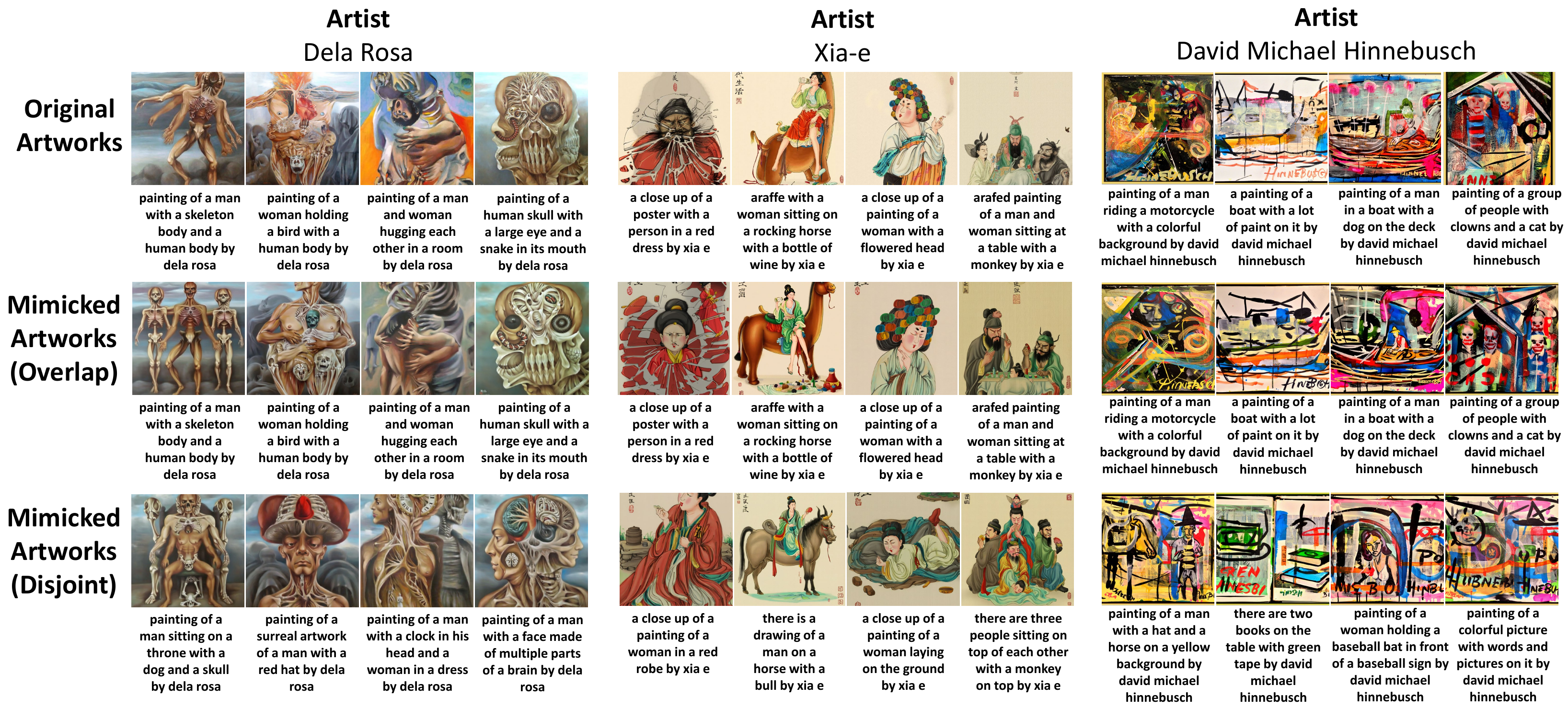}
\caption{The original artworks and mimicked artworks of the online platform \href{https://www.scenario.com/}{Scenario}. The text below the artwork is the corresponding prompt. 
}
\label{fig:senario-images}
\end{figure*}

\mypara{Observations}
We have the following observations from \autoref{tab:performance-in-real-world}. 
1) \sysname achieves the correct auditing results in all experimental settings. 
The auditing results of \sysname on three artists are significantly higher than the threshold 0, which means that \sysname is a valid auditing solution.
For example, ~\autoref{tab:performance-in-real-world-cross-valid} shows the confidence scores calculated by
\sysname in the case of complete overlap.
Confidence scores exceed 0 when the suspicious models undergo fine-tuning with the target artist's artworks, whereas they fall below 0 when the target artist's artworks are not used.
2) \sysname maintains high auditing performance under dataset transfer settings. 
Compared to the auditing results in \autoref{sec:transferability}, \sysname seems to show better dataset transferability on the online platform. 
The reason is mainly that online platforms have better computing power, which makes it possible to get a good artistic imitation even in a disjoint case (please refer to \autoref{fig:senario-images} for the generated images). 

\begin{table}[!t]
        \centering
        \caption{The average of confidence scores predicted by \sysname in the dataset transferring scenarios.
        The results are significantly higher than 0, meaning that \sysname is valid for real-world auditing. 
        } 
        \scalebox{0.8}{
        \begin{tabular}{c|c|c|c}
            \hline
            \diagbox[dir=NW]{\textbf{Artist}}{\textbf{Confidence Score}}{\textbf{Setting}} & \textbf{Completely} & \textbf{Partially} & \textbf{Disjoint} \\
            \hline
            \textbf{Dela Rosa}                & 0.840 & 0.874 & 0.891 \\
            \hline
            \textbf{Xia-e}                    & 0.380 & 0.437 & 0.501 \\
            \hline
            \textbf{David Michael Hinnebusch} & 0.745 & 0.762 & 0.807 \\
            \hline
        \end{tabular}
        }
        \label{tab:performance-in-real-world}
\end{table}

\begin{table}[!t]
        \centering
        \caption{The average of confidence scores predicted by \sysname in cross-validation between three artists. The vertical axis represents the target artist, and the horizontal axis denotes the artist's dataset utilized for the suspicious model's fine-tuning.} 
        \scalebox{0.8}{
        \begin{tabular}{c|c|c|c}
            \hline
            \diagbox[dir=NW]{\textbf{Artist}}{\textbf{Confidence Score}}{\textbf{Artist}} & \textbf{Dela Rosa} & \textbf{Xia-e} & \textbf{D. M. Hinnebusch} \\
            \hline
            \textbf{Dela Rosa}                & 0.866 & -0.836 & -0.858 \\
            \hline
            \textbf{Xia-e}                    & -0.777 & 0.441 & -0.885 \\
            \hline
            \textbf{David Michael Hinnebusch} & -0.924 & -0.772 & 0.776 \\
            \hline
        \end{tabular}
        }
        \label{tab:performance-in-real-world-cross-valid}
\end{table}

\section{Target Models' Performance}
\begin{figure}[!t]
\centering
\includegraphics[width=\hsize]{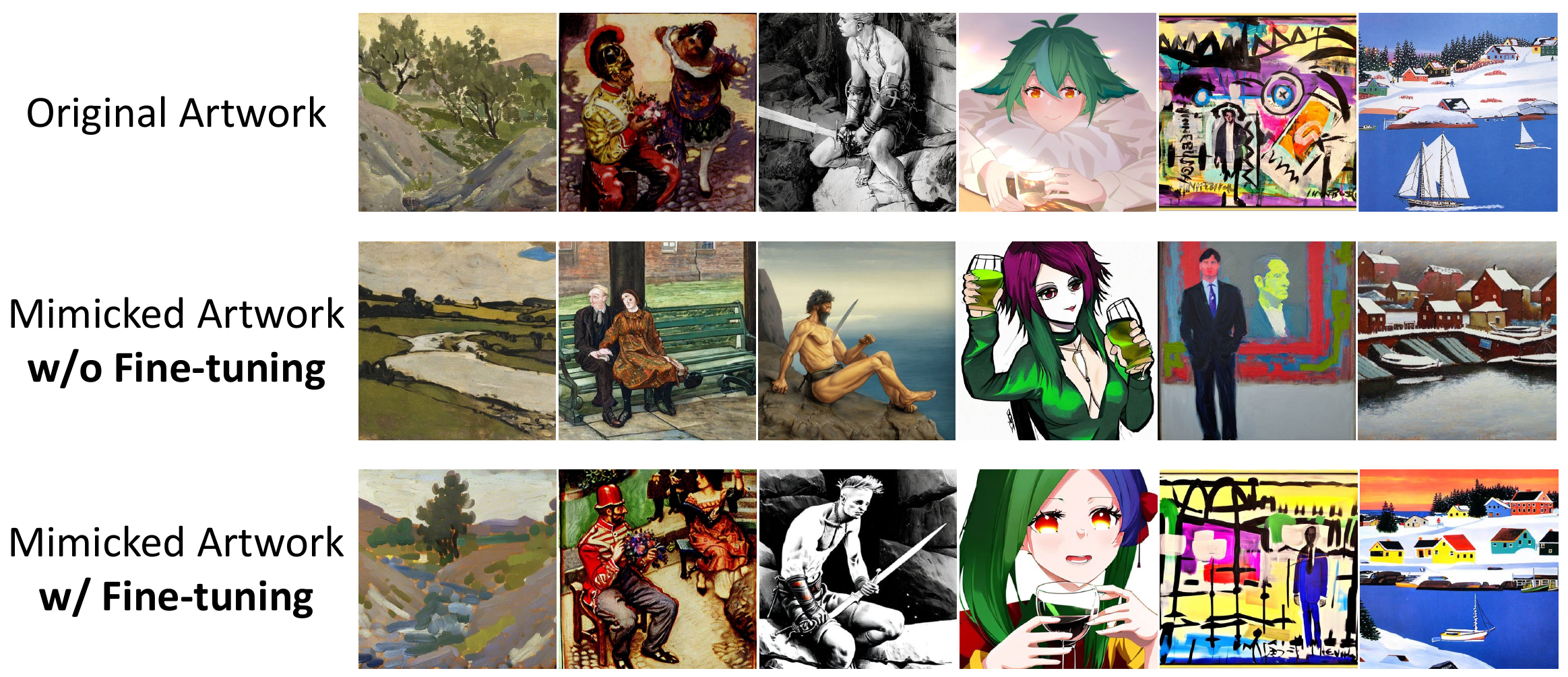}
\caption{Target models' performance. 
The first row displays the original artwork created by the artists.
The second row displays imitations generated by the text-to-image model before its fine-tuning on the original artwork. 
The final row showcases the imitations created after fine-tuning. %
}
\label{fig:finetune-performance}
\end{figure}

We first investigate the stylistic imitation ability of the target model, as shown in \autoref{fig:finetune-performance}. 
The first row shows the original artworks created by artists. 
The second row shows generated artworks without fine-tuning the target models with the original artwork. 
The third row shows mimicked artworks by the target models fine-tuned on the original artworks. 

By comparing these three parts in \autoref{fig:finetune-performance}, it becomes apparent that the target model, after being fine-tuned on the original artworks, exhibits a discernible ability to imitate artistic styles. 
However, detecting the imitation of certain artwork is not immediately evident, making it challenging to ascertain through direct visual inspection, such as the image in the lower left corner of \autoref{fig:finetune-performance}. 
This underscores the necessity of \sysname to identify potential infringements. 

\section{Related Work}
\label{sec:related-work}
In this section, we go into depth about the existing solutions, as the extension of that in \autoref{sec:introduction}. 
As diffusion models continue to evolve and gain popularity, users can now create a vast array of generative works at a low cost, which leads to the negative effects of the replication becoming more acute~\cite{SSGGG22}. 
Especially the artist community is concerned about the copyright infringement of their work~\cite{civitai2022, aigame, MHM2023Finetune}. 
Recently, researchers have proposed a lot of countermeasures to solve this issue~\cite{du2024sok}.

\mypara{Perturbation-based Method}
The artists can introduce slight perturbations that modify the latent representation during the diffusion process, preventing models from generating the expected images. 
Shan~\etal~\cite{shan2023glaze} introduce Glaze, a tool that allows artists to apply ``style cloaks'' to their artwork, introducing subtle perturbations that mislead generative models attempting to replicate a specific artist's style. 
Similarly, Anti-DreamBooth~\cite{Van_Le_2023_ICCV} is a defense system designed to protect against the misuse of DreamBooth by adding slight noise perturbations to images before they are published, thereby degrading the quality of images generated by models trained on these perturbed datasets. 
Chen~\etal~\cite{Chen2023EditShieldPU} propose EditShield, a protection method that introduces imperceptible perturbations to shift the latent representation during the diffusion process, causing models to produce unrealistic images with mismatched subjects. 

However, the goal of adversarial perturbation is to disrupt the learning process of diffusion models, which is orthogonal to the copyright auditing focus of this paper. 
Moreover, adversarial perturbation essentially blocks any legitimate use of subject-driven synthesis based on protected images. 

\begin{table*}[htbp]
\centering
\caption{Dataset Transferability of \sysname. 
``thold'' indicates the threshold-based auditing strategy. 
``t-test'' denotes the hypothesis testing-based auditing strategy. 
}
\scalebox{1.0}{
\begin{tabular}{c|c|ccc|ccc}
\hline
\multirow{2}{*}{\textbf{Model}}     & \textbf{Setting}                                   & \multicolumn{3}{c|}{\textbf{Partially Overlap}}                        & \multicolumn{3}{c}{\textbf{Disjoint}}                                  \\ \cline{2-8} 
                                    & \diagbox[dir=NW]{\textbf{Metric}}{\textbf{Method}} & \textbf{\cite{Pang2023BlackboxMI}} & \textbf{thold}  & \textbf{t-test} & \textbf{\cite{Pang2023BlackboxMI}} & \textbf{thold}  & \textbf{t-test} \\ \hline
\multirow{4}{*}{\textbf{SD-V2}}     & \textbf{Accuracy}                                  & 0.789$\pm$0.042                    & 0.800$\pm$0.021 & 0.760$\pm$0.025 & 0.556$\pm$0.031                    & 0.727$\pm$0.013 & 0.687$\pm$0.016 \\
                                    & \textbf{AUC}                                       & 0.991$\pm$0.007                    & 0.964$\pm$0.008 & /               & 0.699$\pm$0.034                    & 0.956$\pm$0.015 & /               \\
                                    & \textbf{F1 Score}                                  & 0.745$\pm$0.057                    & 0.754$\pm$0.026 & 0.683$\pm$0.043 & 0.281$\pm$0.084                    & 0.623$\pm$0.026 & 0.543$\pm$0.035 \\
                                    & \textbf{FPR}                                       & 0.000$\pm$0.000                    & 0.000$\pm$0.000 & 0.000$\pm$0.000 & 0.000$\pm$0.000                    & 0.000$\pm$0.000 & 0.000$\pm$0.000 \\ \hline
\multirow{4}{*}{\textbf{SDXL}}      & \textbf{Accuracy}                                  & 0.689$\pm$0.031                    & 0.920$\pm$0.027 & 0.873$\pm$0.013 & 0.511$\pm$0.031                    & 0.727$\pm$0.025 & 0.633$\pm$0.021 \\
                                    & \textbf{AUC}                                       & 0.921$\pm$0.012                    & 1.000$\pm$0.000 & /               & 0.872$\pm$0.011                    & 0.980$\pm$0.020 & /               \\
                                    & \textbf{F1 Score}                                  & 0.576$\pm$0.043                    & 0.912$\pm$0.031 & 0.855$\pm$0.017 & 0.148$\pm$0.105                    & 0.622$\pm$0.047 & 0.419$\pm$0.053 \\
                                    & \textbf{FPR}                                       & 0.000$\pm$0.000                    & 0.000$\pm$0.000 & 0.000$\pm$0.000 & 0.000$\pm$0.000                    & 0.000$\pm$0.000 & 0.000$\pm$0.000 \\ \hline
\multirow{4}{*}{\textbf{Kandinsky}} & \textbf{Accuracy}                                  & 0.933$\pm$0.000                    & 0.933$\pm$0.030 & 0.967$\pm$0.000 & 0.711$\pm$0.031                    & 0.907$\pm$0.044 & 0.853$\pm$0.040 \\
                                    & \textbf{AUC}                                       & 0.936$\pm$0.022                    & 0.996$\pm$0.004 & /               & 0.744$\pm$0.017                    & 0.982$\pm$0.013 & /               \\
                                    & \textbf{F1 Score}                                  & 0.923$\pm$0.024                    & 0.938$\pm$0.026 & 0.967$\pm$0.001 & 0.667$\pm$0.067                    & 0.896$\pm$0.055 & 0.826$\pm$0.051 \\
                                    & \textbf{FPR}                                       & 0.187$\pm$0.070                    & 0.133$\pm$0.060 & 0.053$\pm$0.027 & 0.190$\pm$0.067                    & 0.013$\pm$0.027 & 0.000$\pm$0.000 \\ \hline
\end{tabular}
}
\label{table:full-dataset-transferability}
\end{table*}

\begin{table*}[htbp]
\centering
\caption{Model Transferability of \sysname. We use CLIP and BLIP as image captioning models. For each combination, the former is the image captioning model used by the auditor. The later is the image captioning model used in suspicious models. 
} 
\scalebox{0.85}{
\begin{tabular}{c|c|cc|cc|cc|cc}
  \hline
  \multirow{3}{*}{\textbf{Model}} & \textbf{Image Captioning Model} & \multicolumn{2}{c|}{\textbf{CLIP+CLIP}} & \multicolumn{2}{c|}{\textbf{CLIP+BLIP}} & \multicolumn{2}{c|}{\textbf{BLIP+CLIP}} & \multicolumn{2}{c}{\textbf{BLIP+BLIP}} \\
  \cline{2-10}
  ~ & \diagbox[dir=NW]{\textbf{Metric}}{\textbf{Method}} & \textbf{thold} & \textbf{t-test} & \textbf{thold} & \textbf{t-test} & \textbf{thold} & \textbf{t-test} & \textbf{thold} & \textbf{t-test}   \\
  \hline
  \multirow{4}{*}{\textbf{SD-V2}}      & Accuracy            & 0.953$\pm$0.045 & 0.880$\pm$0.045 & 0.853$\pm$0.027 & 0.827$\pm$0.025   & 0.873$\pm$0.025 & 0.807$\pm$0.025 & 0.913$\pm$0.027 & 0.833$\pm$0.021  \\
                              & AUC                 & 0.992$\pm$0.009 & /               & 0.952$\pm$0.011 & /                 & 0.967$\pm$0.007 & /               & 0.972$\pm$0.009 & /                \\ 
                              & F1 Score            & 0.951$\pm$0.049 & 0.864$\pm$0.054 & 0.840$\pm$0.033 & 0.789$\pm$0.036   & 0.859$\pm$0.028 & 0.759$\pm$0.039 & 0.911$\pm$0.026 & 0.806$\pm$0.025  \\
                              & FPR                 & 0.027$\pm$0.033 & 0.013$\pm$0.027 & 0.067$\pm$0.000 & 0.000$\pm$0.000   & 0.027$\pm$0.033 & 0.000$\pm$0.000 & 0.053$\pm$0.050 & 0.027$\pm$0.033  \\
  \hline  
  \multirow{4}{*}{\textbf{SDXL}}       & Accuracy            & 0.947$\pm$0.016 & 0.867$\pm$0.021 & 0.940$\pm$0.025 & 0.873$\pm$0.039   & 0.860$\pm$0.025 & 0.767$\pm$0.037 & 0.900$\pm$0.021 & 0.860$\pm$0.033  \\
                              & AUC                 & 1.000$\pm$0.000 & /               & 1.000$\pm$0.000 & /                 & 0.993$\pm$0.004 & /               & 0.995$\pm$0.007 & /                \\ 
                              & F1 Score            & 0.943$\pm$0.018 & 0.845$\pm$0.028 & 0.935$\pm$0.029 & 0.853$\pm$0.050   & 0.836$\pm$0.033 & 0.693$\pm$0.060 & 0.888$\pm$0.026 & 0.835$\pm$0.046  \\
                              & FPR                 & 0.000$\pm$0.000 & 0.000$\pm$0.000 & 0.000$\pm$0.000 & 0.000$\pm$0.000   & 0.000$\pm$0.000 & 0.000$\pm$0.000 & 0.000$\pm$0.000 & 0.000$\pm$0.000  \\
  \hline  
  \multirow{4}{*}{\textbf{Kandinsky}}  & Accuracy            & 0.933$\pm$0.021 & 0.973$\pm$0.025 & 0.953$\pm$0.016 & 0.967$\pm$0.021   & 0.980$\pm$0.027 & 0.980$\pm$0.016 & 0.987$\pm$0.027 & 0.973$\pm$0.013  \\
                              & AUC                 & 0.998$\pm$0.004 & /               & 0.998$\pm$0.002 & /                 & 1.000$\pm$0.000 & /               & 0.999$\pm$0.002 & /                \\ 
                              & F1 Score            & 0.938$\pm$0.019 & 0.975$\pm$0.023 & 0.956$\pm$0.015 & 0.966$\pm$0.021   & 0.981$\pm$0.025 & 0.979$\pm$0.017 & 0.988$\pm$0.025 & 0.973$\pm$0.014  \\
                              & FPR                 & 0.133$\pm$0.042 & 0.053$\pm$0.050 & 0.093$\pm$0.033 & 0.027$\pm$0.033   & 0.040$\pm$0.053 & 0.000$\pm$0.000 & 0.027$\pm$0.053 & 0.013$\pm$0.027  \\
  \hline
\end{tabular}}
\label{table:full-model-transferability}
\end{table*}

\mypara{Watermark-based Method}
This framework adds subtle watermarks to digital artworks to protect copyrights while preserving the artist's expression. 
Cui~\etal~\cite{Cui2023DiffusionShieldAW} construct the watermark by converting the copyright message into an ASCII-based binary sequence and then translating it into a quaternary sequence. 
During the copyright auditing, they adopt a ResNet-based decoder to recover the watermarks from the images generated by a third-party model. 
Luo~\etal~\cite{Luo2023StealMA} choose to embed subtle watermarks in digital artwork to protect copyrights while preserving the artist's style. 
If used as training data, these watermarks become detectable markers, where the auditor can reveal unauthorized mimicry by analyzing their distribution in generated images. 
Ma~\etal~\cite{Ma2023GenerativeWA} propose GenWatermark, a novel system that jointly trains a watermark generator and detector. 
By integrating the subject-driven synthesis process during training, GenWatermark fine-tunes the detector with synthesized images, boosting detection accuracy, and ensuring subject-specific watermark uniqueness.
Zheng~\etal~\cite{ZXPLRCCX24} introduce TabularMark, a watermarking scheme based on hypothesis testing.
They employ data noise partitioning for embedding, allowing adaptable perturbation of both numerical and categorical attributes without compromising data utility.

However, given that digital artworks are already in the public domain, artists must utilize a post-publication mechanism that does not depend on the prior insertion of altered samples into the dataset. 
In contrast, watermarking constitutes a preemptive measure, necessitating the integration of manipulated samples into the dataset before its release. 

\begin{table*}[htbp]
\centering
\caption{Impact of data augmentation and distortion calibration. ``w/o DA'' shows the auditing performance without data augmentation. ``w/o DC'' shows the auditing performance without distortion calibration. } 
\scalebox{1.0}{
\begin{tabular}{c|c|cc|cc|cc}
\hline
\multirow{2}{*}{\textbf{Model}}     & \textbf{Setting} & \multicolumn{2}{c|}{\textbf{w/o Data Augmentation}} & \multicolumn{2}{c|}{\textbf{w/o Distortion Calibration}} & \multicolumn{2}{c}{\textbf{Baseline}} \\ \cline{2-8} 
                                    & \textbf{Method}  & \textbf{thold}           & \textbf{t-test}          & \textbf{thold}              & \textbf{t-test}            & \textbf{thold}    & \textbf{t-test}   \\ \hline
\multirow{4}{*}{\textbf{SD-V2}}     & Accuracy         & 0.927$\pm$0.025          & 0.867$\pm$0.021          & 0.953$\pm$0.016             & 0.853$\pm$0.045            & 0.953$\pm$0.045   & 0.880$\pm$0.045   \\
                                    & AUC              & 0.995$\pm$0.005          & /                        & 0.994$\pm$0.008             & /                          & 0.992$\pm$0.009   & /                 \\
                                    & F1 Score         & 0.920$\pm$0.029          & 0.845$\pm$0.028          & 0.951$\pm$0.018             & 0.825$\pm$0.060            & 0.951$\pm$0.049   & 0.864$\pm$0.054   \\
                                    & FPR              & 0.000$\pm$0.000          & 0.000$\pm$0.000          & 0.013$\pm$0.027             & 0.000$\pm$0.000            & 0.027$\pm$0.033   & 0.013$\pm$0.027   \\ \hline
\multirow{4}{*}{\textbf{SDXL}}      & Accuracy         & 0.633$\pm$0.052          & 0.620$\pm$0.062          & 0.953$\pm$0.016             & 0.893$\pm$0.033            & 0.947$\pm$0.016   & 0.867$\pm$0.021   \\
                                    & AUC              & 0.874$\pm$0.069          & /                        & 0.997$\pm$0.002             & /                          & 1.000$\pm$0.000   & /                 \\
                                    & F1 Score         & 0.411$\pm$0.117          & 0.372$\pm$0.149          & 0.951$\pm$0.018             & 0.879$\pm$0.042            & 0.943$\pm$0.018   & 0.845$\pm$0.028   \\
                                    & FPR              & 0.000$\pm$0.000          & 0.000$\pm$0.000          & 0.000$\pm$0.000             & 0.000$\pm$0.000            & 0.000$\pm$0.000   & 0.000$\pm$0.000   \\ \hline
\multirow{4}{*}{\textbf{Kandinsky}} & Accuracy         & 0.647$\pm$0.027          & 0.620$\pm$0.034          & 0.880$\pm$0.016             & 0.913$\pm$0.016            & 0.933$\pm$0.021   & 0.973$\pm$0.025   \\
                                    & AUC              & 0.850$\pm$0.085          & /                        & 0.977$\pm$0.017             & /                          & 0.998$\pm$0.004   & /                 \\
                                    & F1 Score         & 0.460$\pm$0.075          & 0.382$\pm$0.090          & 0.893$\pm$0.013             & 0.920$\pm$0.014            & 0.938$\pm$0.019   & 0.975$\pm$0.023   \\
                                    & FPR              & 0.013$\pm$0.027          & 0.000$\pm$0.000          & 0.240$\pm$0.033             & 0.173$\pm$0.033            & 0.133$\pm$0.042   & 0.053$\pm$0.050   \\ \hline
\end{tabular}}
\label{table:full-DA-and-DC-impact}
\end{table*}

\end{document}